\documentclass[10pt,journal,compsoc]{IEEEtran}

	\ifCLASSOPTIONcompsoc
	  \usepackage[nocompress]{cite}
	\else
	  \usepackage{cite}
	\fi

	\ifCLASSINFOpdf
	  \usepackage[pdftex]{graphicx}
	\else
	\fi

	\usepackage{amsmath}

	\usepackage{array}

	\usepackage{graphicx}
    \usepackage{bm}
	\usepackage{url}
	\usepackage{color}
	\usepackage{multirow}
	\usepackage{booktabs} 
	\usepackage{floatrow}
	\usepackage{amssymb}
	\usepackage{threeparttable}
	\usepackage{geometry}
	\newfloatcommand{capbtabbox}{table}[][\FBwidth]
	\newcommand{\tabincell}[2]{\begin{tabular}{@{}#1@{}}#2\end{tabular}}
	
	\newcommand{\qz}{\textcolor[rgb]{0,0,0}}
	
	\newcommand{\ws}{\textcolor[rgb]{0,0,0}}
	\newcommand{\wss}{\textcolor[rgb]{0,0,0}}
	\newcommand{\qqz}{\textcolor[rgb]{0,0,0}}
	\newcommand{\qzz}{\textcolor[rgb]{0,0,0}}
	
	\newcommand{\newc}{\textcolor[rgb]{0,0,0}}
	\newcommand{\chgorder}{\textcolor[rgb]{0,0,0}}
	\newcommand{\nnc}{\textcolor[rgb]{0,0,0}}
	\newcommand{\nncc}{\textcolor[rgb]{0,0,0}}
	\newcommand{\lastc}{\textcolor[rgb]{0,0,0}}
	\newcommand{\lastcc}{\textcolor[rgb]{0,0,0}}
	\newcommand{\revision}{\textcolor[rgb]{0,0,0}}
	\newcommand{\rchange}{\textcolor[rgb]{0,0,0}}
	\newcommand{\roundtwo}{\textcolor[rgb]{0,0,0}}
\begin{document}

\newgeometry{top=6cm,bottom=1cm}

\onecolumn{

\noindent \textbf{\huge{Person Re-identification by Contour Sketch under Moderate Clothing Change}}

\vspace{2cm}

\noindent {\LARGE{Qize Yang, Ancong Wu, Wei-Shi Zheng}}

\Large
\vspace{2cm}

\noindent Our dataset is available at:

\noindent \url{http://www.isee-ai.cn/%7Eyangqize/clothing.html}

\vspace{1cm}
\noindent For reference of this work, please cite:

\vspace{1cm}
\noindent Qize Yang, Ancong Wu, Wei-Shi Zheng.
``Person Re-identification by Contour Sketch under Moderate Clothing Change.''
\emph{IEEE Transactions on Pattern Analysis and Machine Intelligence.} (DOI 10.1109/TPAMI.2019.2960509).

\vspace{1cm}

\noindent Bib:\\
\noindent
@article\{yang2020person,\\
\ \ \   title=\{Person Re-identification by Contour Sketch under Moderate Clothing Change\},\\
\ \ \  author=\{Yang, Qize and Wu, Ancong and Zheng, Wei-Shi\},\\
\ \ \  journal=\{IEEE Transactions on Pattern Analysis and Machine Intelligence\\(DOI 10.1109/TPAMI.2019.2960509)\},\\
\ \ \  year=\{2020\}, \\
\}

}

\clearpage

\newpage
\restoregeometry

%
\title{Person Re-identification by Contour Sketch under Moderate Clothing Change}

\author{Qize~Yang,
	Ancong~Wu,
	Wei-Shi~Zheng
	\IEEEcompsocitemizethanks{
		\IEEEcompsocthanksitem Qize Yang is with the School of Data and Computer Science, Sun Yat-sen University, Guangzhou 510275, China. E-mail: yangqz@mail2.sysu.edu.cn.
		\IEEEcompsocthanksitem Ancong Wu is with the School of Electronics and Information Technology, Sun Yat-sen University, Guangzhou 510275, China, and is also with Guangdong
		Province Key Laboratory of Information Security, China. E-mail: wuancong@mail2.sysu.edu.cn.
		\IEEEcompsocthanksitem Wei-Shi Zheng is with the School of Data and Computer Science, Sun Yat-sen University, Guangzhou 510275, China, 
		with Peng Cheng Laboratory, Shenzhen 518005, China,
		and is also with the Key Laboratory of Machine Intelligence and Advanced Computing (Sun Yat-sen University), Ministry of Education, China. 
E-mail: wszheng@ieee.org /zhwshi@mail.sysu.edu.cn.
\protect\\ (Corresponding author: Wei-Shi Zheng.)}
}

%
%

\markboth{CITATION INFORMATION: DOI 10.1109/TPAMI.2019.2960509, IEEE TRANSACTIONS ON PATTERN ANALYSIS AND MACHINE INTELLIGENCE}%
{Shell \MakeLowercase{\textit{et al.}}: Bare Demo of IEEEtran.cls for Computer Society Journals}
%



\IEEEtitleabstractindextext{%
	\begin{abstract}
		Person re-identification (re-id), the process of matching pedestrian images across different camera views, is an important task in visual surveillance. Substantial development of re-id has recently been observed, and the majority of existing models are largely dependent on color appearance and assume that pedestrians do not change their clothes across camera views. 
		\ws{This limitation, however, can be an issue for re-id when tracking a person at different places and at different time if that person (e.g., a criminal suspect) changes his/her clothes, causing most existing methods to fail, since they are heavily relying on \nncc{color} appearance and thus they are inclined to match a person to another person wearing similar clothes. In this work, we call the person re-id under clothing change the ``cross-clothes	person re-id''.
		In particular, we consider the case when a person only changes his clothes moderately as a first attempt at solving this problem based on visible light images; that is we assume that a person wears clothes of a similar thickness, and thus the shape of a person would not change significantly when the weather does not change substantially within a short period of time.
		We perform \ws{cross-clothes person re-id} based on a contour sketch of person image to take advantage of the shape of the human body instead of color information for extracting features that are robust to \ws{moderate} clothing change.} 
		To select/sample more reliable and discriminative curve patterns on a body contour sketch, we introduce a learning-based spatial polar transformation (SPT) layer in the deep neural network to transform \nnc{contour} sketch images for extracting reliable and discriminant convolutional neural network (CNN) features in a polar coordinate space. 
		An angle-specific extractor (ASE) is applied in the following layers to extract more fine-grained discriminant \qz{angle}-specific features. \newc{By varying the sampling range of the SPT, we develop a multistream network for aggregating multi-granularity} features to better identify a person.
		Due to the lack of a large-scale dataset for cross-clothes person re-id, we contribute a new dataset that consists of 33698 images from 221 identities. Our experiments illustrate the challenges of cross-clothes person re-id and demonstrate the effectiveness of our proposed method.
	\end{abstract}

	}

\maketitle

\IEEEdisplaynontitleabstractindextext

%
\IEEEpeerreviewmaketitle

\IEEEraisesectionheading{\section{Introduction}\label{sec:introduction}}

%
%
%
%
Person re-identification (re-id) is the process of associating a single person
who moves across disjoint camera views. Re-id is becoming more
popular as intelligent video surveillance becomes increasingly
important.
The development of re-id ranges from extracting features \cite{ojala1996comparative,liao2015person,dalal2005histograms,gray2008viewpoint,hirzer2012relaxed} and distance metric learning
\cite{liao2015person,li2013learning,koestinger2012large,zheng2012reidentification,kviatkovsky2013color,
yu2018unsupervised,chen2018person,zheng2015towards}
to deep-learning-based methods
\cite{ahmed2015improved,li2014deepreid,xiao2016learning,yang2014salient,xu2017jointly,liu2017hydraplus,sun2017beyond,wu2020rgbir,yin2020fine}.
These methods are designed mainly to overcome changes
in viewing angle, background clutter, body pose, scale, and occlusion.
The performance of re-id methods has recently improved rapidly
by adopting deep features and by using metric learning methods, semantic
attributes and appearance models.

\begin{figure*}[t]
	\centering
	\includegraphics[height=6.5cm,trim=0 0 0 0,clip]{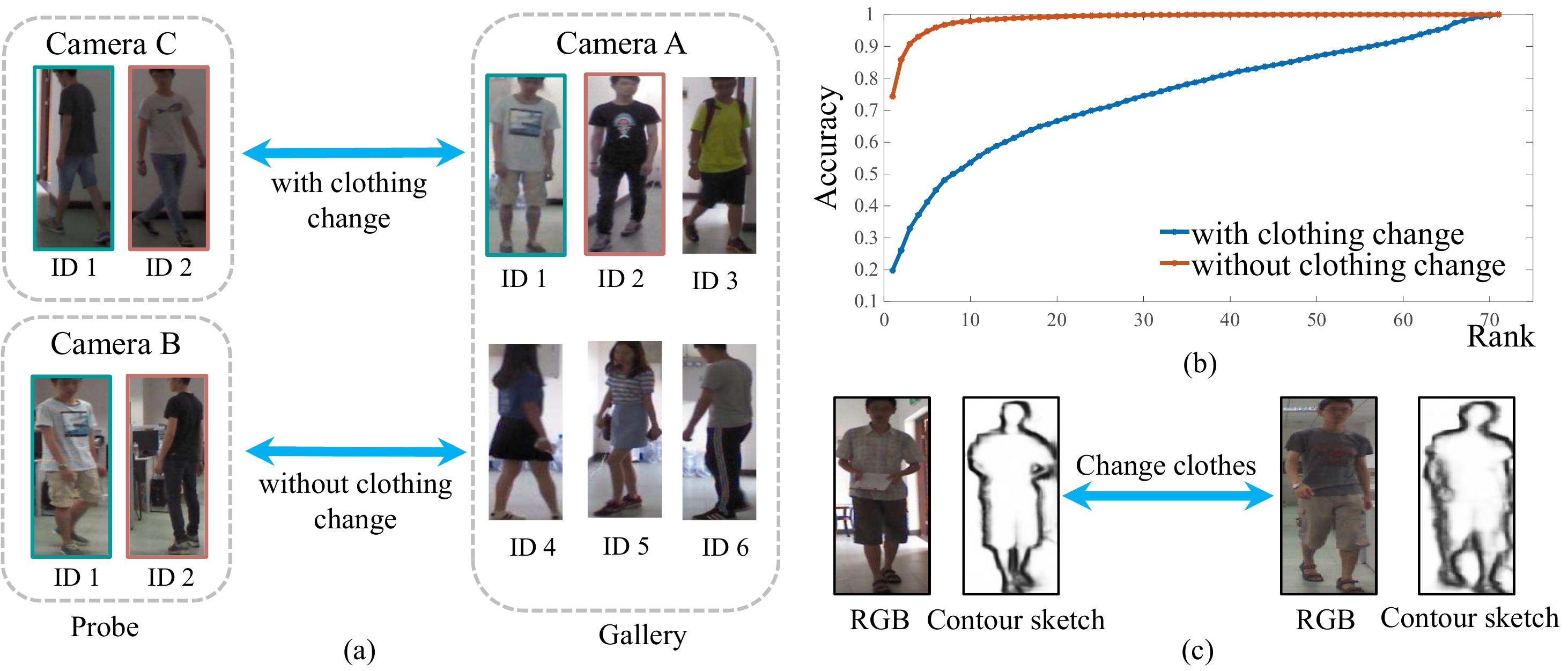}
	\caption{(a) Illustration of person re-id.
		The matching between Camera A and Camera C is cross-clothes, whereas the matching between Camera A and
		Camera B is without clothing changes.
		(b) CMC curve of the result of person re-id. The blue curve indicates matching
		with clothing change, whereas the orange curve shows the
		result of the matching without clothing change.
		(c) RGB and contour sketch images of the same person in different clothes.
	}
	\label{fig:illustration}
\end{figure*}

Existing state-of-the-art person re-id models assume that a person does not change his/her clothes when he/she appears in different camera views within a short period.
However, a person could put on or take off some clothes (e.g., due to weather changes) at different places and at different time, or more critically, a criminal might try to hide himself/herself by randomly changing his/her appearance; thus, existing re-id systems and even humans face challenges when attempting to re-id such a person at a distance. In a real case that occurred in China, 1000 policemen watched nearly 300 TB \revision{(about 172800 hours for videos in $1920\times 1080$ resolution, 25 fps, under the H.264 video compression standard)} of surveillance videos in just two months to capture a serial killer who was very skillful at anti-tracking and often changed his clothes. However, the special body shape of the murderer, a unique characteristic, was one of the important cues that helped to finally trace him \cite{zhou_wiki,zhou_bbc,zhou_china}.


Unlike other variations, such as view change and lighting changes, in cross-view matching,  clothing changes are essentially impossible to model.
For example, as shown in Figure~\ref{fig:illustration} (a), two people appearing in Camera \qzz{C} are wearing different clothes in Camera \qzz{A}. In such a case, color information is unreliable or even misleading. As shown in Figure~\ref{fig:illustration} (b), a test result of Res-Net50 \cite{he2016deep} on our collected dataset, which is introduced later, shows a sharp performance drop between the cases of a person wearing the same clothes and wearing different clothes. In this work, we call the process of person re-id under clothing change the ``cross-clothes person re-id'' problem.

We consider cross-clothes person re-id under moderate clothing changes as a first attempt at solving this problem based on visible light images without using other devices (such as depth sensors); that is, we assume that people wear clothes of a similar thickness and, thus,  that the shape of a person would not change significantly when the weather does not change substantially within a short period of time. This assumption is empirically justified in Figure~\ref{fig:illustration} (c), which shows that the shape of a person is very similar when he/she wears different clothes of similar thickness.

On the basis of the above assumption, we attempt to investigate the use of contour sketches to perform person re-id under clothing change. Compared to conventional person re-id, the intraclass variation can be large or unpredictable. In such a context, we argue that compared to color-based visual cues, the contour sketch provides reliable and effective visual cues for overcoming the discrepancy
between images of the same person. However, the application of a contour sketch extractor for solving this problem is not straightforward.
We find that the \newc{contour sketches of different human bodies look similar globally} and, more importantly, not all the curve patterns on the contour \newc{sketch} are discriminative. Therefore, we develop a learning-based \textbf{s}patial \textbf{p}olar \textbf{t}ransformation (SPT) to automatically select/sample relatively invariant, reliable and discriminative local curve patterns. Additionally, we introduce an \textbf{a}ngle \textbf{s}pecific \textbf{e}xtractor (ASE) to model the interdependencies between the channels of the feature map for each angle stripe to explore the fine-grained \qz{angle}-specific features. Finally, a multistream network is learned for an \qz{ensemble} of features to extract \newc{multi-granularity (i.e. global coarse-grained and local fine-grained)} features and to re-id a person when he/she dresses differently.


We contribute a new person re-id dataset named Person Re-id under moderate Clothing Change (PRCC) to study person re-id under clothing change. This dataset contains 33698 images from 221 people, captured by 3 cameras,	as shown in Figure~\ref{fig:illustration} (a). Each person wears the same clothes in Camera A and Camera B but different clothes in Camera C. Our experiments on the PRCC dataset illustrate the challenge of cross-clothes person re-identification, and we find that not only \qqz{hand-crafted} features but also popular deep-learning-based methods achieve unsatisfactory performance in this case. In comparison, we show that our proposed method achieves the highest accuracy for person re-id under clothing change.

The main contributions of this work are summarized as follows:
\begin{itemize}
	\item We find that for person re-id under moderate clothing change, contour sketches are much more effective than the conventional person re-id methods based on color visual cues.
	\item We design a new \ws{deep} {contour-sketch-based} network \ws{for overcoming} person re-id under moderate clothing changes. Specifically, to quantify the contour sketch images effectively, we introduce SPT to select the relatively invariant and discriminative contour patterns and ASE to explore \qz{angle}-specific fine-grained discriminant features.
	\item  We contribute a new person re-id dataset with moderate clothing changes, namely, the PRCC dataset.
\end{itemize}
\nncc{
In our experiments,
	      we not only quantitatively analyze the challenge of the clothing change problem by varying the degree of clothing change but also analyze the performance of person re-id when clothing changes are combined with other challenges.}

\begin{figure*}[t]
	\centering
	\includegraphics[height=5.5cm,trim=0 0 0 0,clip]{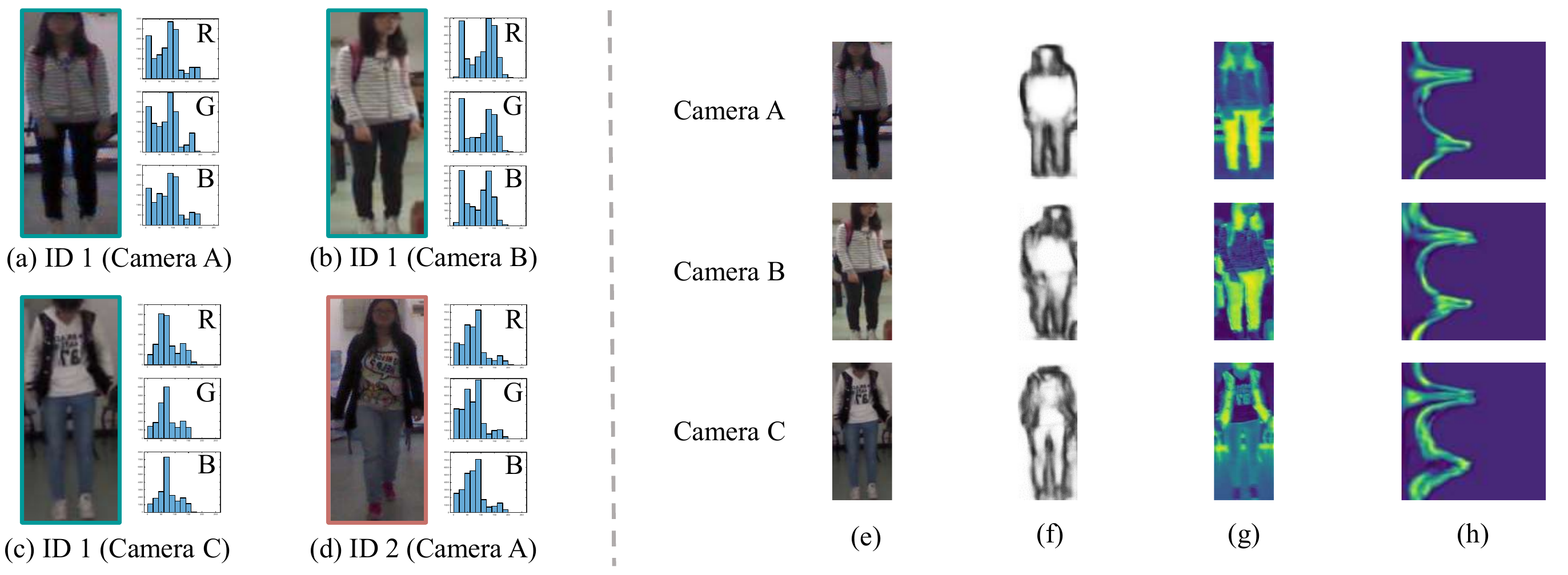}
	\caption{
		(a)-(d) RGB histograms of samples in different cases.
		When the person with ID 1 wears different clothes,
		the appearance information in the RGB histogram changes substantially (see (a) and (c))
		compared to the case of wearing the same clothes (see (a) and (b)).
		The appearance even becomes confusing compared to that of the person with ID 2 in similar clothes
		(see (c) and (d)). In this case, the color information can be unreliable and
		misleading. \wss{(e) and (f) demonstrate some examples of input RGB images and contour sketch images captured in three camera views of the PRCC testing set, respectively. (g) and (h) are the corresponding visualizations of the output feature maps of the first convolution block of Res-Net50 and our model (i.e. with SPT (see text in Section \ref{sec:spt})), respectively. In comparison, the proposed SPT features are more stable for the same person under clothing change.
		}
	}
	\label{fig:problem}
\end{figure*}

\section{Related Works}

\subsection{Person Re-identification}
Reliable feature representation for each person image is anticipated due to various visual changes in person images across camera views.
Representative descriptors, such as LBP \cite{ojala1996comparative}, LOMO\cite{liao2015person}, HOG \cite{dalal2005histograms} and
BoW \cite{zheng2009associating,ma2012local}, are designed to extract texture and color features.
To make the extracted features more robust against cross-view changes, metric-learning-based re-id models, such as XQDA \cite{liao2015person}, KISSME \cite{koestinger2012large}, RDC \cite{zheng2013reidentification},
PCCA \cite{mignon2012pcca}, LFDA \cite{li2013learning},
DNS \cite{zhang2016learning}, SCSP \cite{chen2016similarity}, and DVAML \cite{chen2018deep}
have been developed to minimize the gap between visual features for cross-view
matching. 
Deep learning models for person re-id \cite{ahmed2015improved,li2014deepreid,chen2017person,xiao2016learning,yang2014salient,xu2017jointly,liu2017hydraplus,song2018mask,xu2018attention,sun2017beyond,suh2018part,yuan2018safenet} have also been developed recently to combine feature learning and metric learning.

However, the above models rely substantially on color information, and in cross-clothes matching problems, color information is unreliable since the texture and color information of a person changes significantly under clothing change. Compared to color cues, the contour and \nncc{shape} are \newc{more} consistent under (moderate) clothing change.





Depth-based re-id methods \cite{munaro2014one,lorenzo2012study,haque2016recurrent,wu2017robust,karianakis2018reinforced} have recently been proposed to
overcome illumination and color changes.
Depth-based methods can partially solve the clothing change problem for person re-id due to the captured 3D human shape; however, depth image capture requires an additional device that is not widely deployed.

We propose using contour sketch images generated
from RGB images instead of raw RGB images as the input of
a re-id model to address the moderate clothing change problem.
Similar to depth-based methods, our contour-sketch-based method utilizes human shape information, but
the difference is that our proposed method \nncc{only} requires estimation of the 2D shape of a human from an RGB image.
While direct consideration of the contour sketch images using CNN is ineffective, we transform the sketch
images using the proposed SPT and a multistream model for extracting selective \newc{multi-granularity} reliable features.

We note that Lu et al. \cite{lu2018cross} proposed a cross-domain adversarial feature learning method for sketch re-id and contributed a sketch person re-id dataset. This work differs from ours notably. First, Lu et al. do not consider solving the clothing change problem in re-id, and they focus on cross-domain matching (i.e., using a sketch image to match the RGB image). Second, the sketch images are painted by professional sketch artists \nncc{in \cite{lu2018cross}}, and some clothing information remains, whereas our contour sketch images for person re-id are generated by an edge detector from the RGB images, representing the human contour.


\subsection{Research on Sketch Image Retrieval}

Our work is related to existing sketch-based retrieval research \cite{yu2016sketch,song2017deep,zhang2016sketchnet}. These studies consider cross-modality search (i.e., using sketch images to retrieve the corresponding RGB images), which is different from our cross-clothes person re-id. Our study focuses on matching the contour sketch images of persons captured with different camera views. Thus, the models for sketch-based retrieval are not optimal for the cross-clothes person re-id problem studied in this paper.

\subsection{Gait Recognition}

The contour explored by our sketch method can be similarly explored by gait recognition methods	to some extent \cite{han2006individual, wang2003silhouette,muramatsu2015gait,wu2017comprehensive,makihara2017joint}, which is also clothing independent. However, these methods make a stringent assumption on the image sequences, that is, the sequences must include a complete gait cycle, making them difficult to apply in ordinary image-based person re-id scenarios.
Our contour-sketch-based method focuses on extracting invariant local spatial cues from still images. By contrast, gait recognition focuses on temporal-spatial motion cues in video sequences, which cannot be applied to still images.

\section{Approach}

\subsection{Problem Statement and Challenges}

For person re-id, the target person may change clothes even in the short term. Commonly, in surveillance for security, criminal suspects disguise themselves by changing clothes and covering their faces with masks to prevent capture. Unreliable clothing and face information is a key challenge for security.

Clothing changes make the re-id task more challenging.
As shown in Figure~\ref{fig:problem}, the histograms of (a) are similar to that of (b) because the people in these images are wearing the same clothes. The histograms of image (c), from the same person wearing different clothes, are clearly different from those of image (a) and image (b). However, the histograms of image (c) are similar to the histograms of (d), which show another person wearing similar clothes. Since the histograms change considerably, this problem is difficult to \qzz{be solved} directly based on appearance information. To further illustrate the challenges, we trained a Res-Net50 \cite{he2016deep} on our dataset. The method worked well for person re-id without clothing changes, reaching \textbf{74.80\%} rank-1 accuracy. By contrast, the model performed poorly when people changed their clothes, with the rank-1 accuracy substantially reduced to \textbf{19.43\%}.

Shape information
is an important cue for recognizing a person when he/she changes his/her clothes and when face information is not reliable due to distance. As shown in Figure~\ref{fig:illustration} (c), the \nnc{contour} sketch images are more consistent than RGB images when a person changes to clothes of a similar thickness. Since the color information is not robust to clothing changes, we consider contour-sketch-based person re-id when shape-invariant features are mined from contour sketch images. In developing this approach, we assume that a person does not change his/her clothes dramatically, for instance, from a T-shirt to a down jacket, as we assume that people would not usually make dramatic changes in clothing when the temperature of the environment does not change considerably.






\vspace{0.1cm}

\noindent \textbf{Challenge of Quantifying Human Contour}. The contour-sketch-based method is potentially an appropriate way to solve the moderate clothing-change problem. 
\qzz{However, \nncc{as shown in our experiments,} the application of a contour extractor is not straightforward because the contour sketches of different human bodies look similar globally, and not all local curve patterns on the contour are discriminative. Therefore, it is demanded to enlarge \newc{the} interclass gap between the contour sketches of different people by exploiting discriminative local feature on the contour sketch.}


To solve the aforementioned challenges, we \newc{design} a \newc{learnable} spatial polar transformation (SPT) to select discriminative curve patterns. We also \nncc{develop} an angle-specific
extractor (ASE) to select robust and discriminative fine-grained features. Furthermore, by developing a multiple-stream framework, our proposed transformation can be extended to extract \newc{multi-granularity} features. Finally, a cross-entropy loss and a triplet margin loss are adopted as the target functions of our multistream network to mine more discriminant clothing-change-invariant features, which increases the interclass gap to differentiate people.

\subsection{Learning Spatial Polar Transformation (SPT) in Deep Neural Network}\label{sec:spt}

\nncc{In order to develop an effective method to select discriminant curve patterns, we aim to seek a transformation $\mathcal{T}$ that transforms a contour sketch image $\bm{U}\in \mathbb{R}^{H\times W}$ into
another shape $\bm{V}\in \mathbb{R}^{N\times M}$.
In this work, we model the transformation $\mathcal{T}$ as a learnable polar transformation,}
\begin{align}
	\label{eq:trans}
	\bm{V}=\mathcal{T}(\bm{U}) \ ,
\end{align}
so as to transform a contour sketch image into polar coordinate to enhance the rotation and scale invariance of contour sketch images \cite{wolberg2000robust,zokai2005image}, as shown in Figure~\ref{fig:rst} (a). By such a transformation, we \nncc{expect to} select the discriminant curve patterns based on its polar angle.



\vspace{0.1cm}

\subsubsection{Transformation by \qz{differentiable polar transformation}}

\begin{figure*}[t]
	\centering
	\includegraphics[height=6.8cm,trim=0 0 0 0,clip]{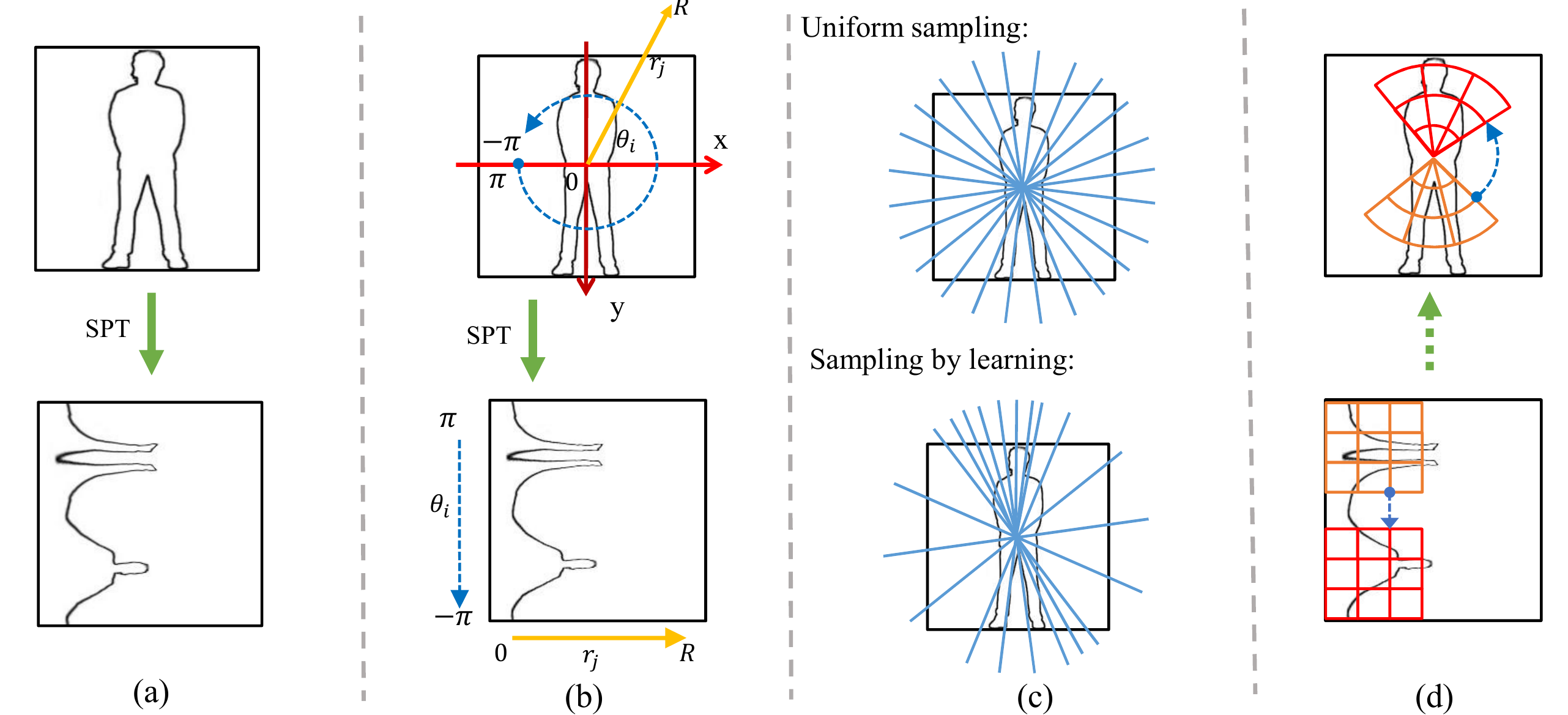}
	\caption{Illustration of spatial polar transformation.
		(a) \qz{Example of our proposed \qzz{spatial polar transformation (SPT).}}
		(b) \qz{The detailed illustration of our proposed transformation,
		where $r_j$ is the radius of the \nnc{contour} sketch image in polar coordinates and
		$\theta_i$ is the polar angle.}
		(c) Uniform sampling method (top) and
		the sampling method for learning \qqz{$\bm\theta$} by SPT (bottom).
		(d) Change in the shape of the receptive field after the transformation.
	}
	\label{fig:rst}
\end{figure*}

The horizontal and vertical axes of images are often represented as the X-axis and Y-axis in Cartesian coordinates in the image processing field, so we can define the position of each pixel in an image using a coordinate pair $(x, y)$. The conversion formula between Cartesian coordinates $(x, y)$ and polar coordinates $(r, \varphi)$ can be written as follows
\begin{align}
	\label{eq:relation}
	r = \sqrt{x^2+y^2}, \varphi=\arctan{\frac{y}{x}}.
\end{align}
By using the angular axis and the radius axis as the vertical axis and horizontal axis of an image, the original image (e.g., the upper part of Figure~\ref{fig:rst} (a))  is transformed into another representation (i.e., the lower part of Figure~\ref{fig:rst} (a)). 

\qzz{The differentiable polar transformation includes two steps, i.e., computing the sampling grid and \newc{performing} sampling. The sampling grid is composed of the pair of sampled angle and sampled radius.} We let $\theta_{i}$ be the $i$-th sampled angle of the \nnc{contour} sketch image,
i.e., the angle of the pixels in polar coordinates, ranging from $\pi$ to $-\pi$ uniformly, where $i=0,\dots,N$. Let $r_j = j\times R/M$ represent the sampled radius in polar coordinates, where $j=0,\dots,M$ and $R$ is the maximum sampled radius.
Based on the sampled angle and sampled radius, we generate the sampling grid by
\begin{align}
	x_{i,j}^s=r_j\cos \theta_i, \
	y_{i,j}^s=r_j\sin \theta_i,
\end{align}
where $x_{i,j}^s$ and $y_{i,j}^s$ represent	the coordinates of the original \nnc{contour} sketch image, \ws{and $i,j$ represent the $i$-{th} row and \qzz{the} $j$-{th} column of the transformed image ($V\in \mathbb{R}^{N\times M}$), respectively.} This formula is similar to the relation between Cartesian coordinates and the polar coordinates \qzz{(i.e. Eq.(\ref{eq:relation}))}.

After generating the sampling grid, the next step is to perform sampling on the \nnc{contour} sketch image with interpolation. Let $v_{i,j}$ be the pixel value of the transformed image; we can use a \qz{differentiable} bilinear sampling kernel \qzz{\cite{jaderberg2015spatial}} to generate the transformed image, that is,
\qz{
	\begin{equation}
		\begin{aligned}
			v_{i,j}= \sum_{h}^{H} \sum_{w}^{W}u_{h,w} \times
			\left \lfloor   1-\left | x_{i,j}^s-w \right | \right \rfloor_+\times
			\left \lfloor  1-\left | y_{i,j}^s-h \right | \right \rfloor_+,
		\end{aligned}
	\end{equation}
	where $\left \lfloor \cdot \right \rfloor_+$ means $\max\{0, \cdot \}$,  and $u_{h,w}$ is the pixel value of the \nnc{contour} sketch image $U\in \mathbb{R}^{H\times W}$. Figure~\ref{fig:rst} (b) presents an intuitive illustration of this transformation.
}

\vspace{0.1cm}

\subsubsection{Learnable spatial polar transformation}
\label{sec:lspt}
Our aim is not only to perform a spatial transformation \qzz{by uniform sampling} and then use  CNN; since different contour parts may contribute differently, we aim to learn a Spatial Polar Transformation (SPT), so that only a portion of the contour
parts are automatically selected/sampled for the spatial transformation.
We use the neural network to automatically learn the sampled angle instead of fixing its value. In this way, we focus more attention on discriminative curve patterns.
Therefore, the curve patterns are unlikely to be uniformly sampled
(see the upper part of Figure 3 (c)) on the contour sketch for the spatial transformation,
and only selective curve patterns are sampled (lower part of Figure 3 (c)).

Our objective is to learn a transformation $\mathcal{T}$ for modeling $V$ with parameters $\bm{\theta}=[\theta_1,\dots,\theta_N]$ \qzz{(i.e. the sampled angles)};  then, we can rewrite Eq.(\ref{eq:trans}) as
\begin{align}
	\bm{V}=\mathcal{T}(\bm{U};\bm{\theta}) \ .
\end{align}
In the deep neural network, we update the parameters by computing the back-propagation gradient and use SGD \cite{sutskever2013importance} or another algorithm to optimize the parameters with respect to the target function. If we use SGD to update $\bm{\theta}$ directly, then the sampled angle is updated without considering \newc{the range of the sampled angle
and the order of the sampled angle (i.e., $\theta_{i+1} \leq \theta_{i}$ in our polar coordinate) is disrupted.} However, the order of the sampled angle retains the semantic structure of the human, which is important for modeling the contour of the human. To maintain the sampled angle $\theta_i$ within a specific range \newc{and the order of sampled angle},  we parametrize $\theta_{i}$ by
\begin{align}
	\theta_{i}=f\left(z_{i}\right)\ ,
	z_{i}=\frac{\sum_{k=0}^{i}  \left \lfloor\lambda_{k} \right \rfloor_+}
	{\sum_{k=0}^{N}\left \lfloor \lambda_{k} \right \rfloor_+},
\end{align}
where $\lambda_k$ is the parameter of the SPT transformation.
$f$ is a linear function that is designed to map the intermediate variable $z_{i}$
to a specific range of $\theta_i$, and in this work, $f$ has the following form:
\begin{align}
	\label{eq:mapping}
	f(z_i)=(b - a) \times z_i + a.
\end{align}
Hence, the sampled angles are computed by parameter $\lambda_i$, so we can constrain $\theta_i$ within a range from $a$ to $b$
\newc{by $f$}, and jointly consider the relation between sampled angles implicitly \newc{by mapping $\lambda_k$ to $z_i$}. \qzz{In this work, we initialize all $\lambda_k$ as a constant number at the beginning, so the sampled angles are uniformly distributed initially; then update the $\lambda_k$ to compute the $\theta_i$ during training.}

\vspace{0.1cm}

\subsubsection{Learning $\bm{\theta}$}

During the backward propagation, the
gradient of $v_{i, j}$ with respect to $x_{i,j}^s$ is
\qz{
	\begin{equation}
		\begin{aligned}
			  & \frac{\partial v_{i,j}}{\partial  x_{i,j}^s}=  \sum_{h}^{H} \sum_{w}^{W}u_{h,w}
			\max \left ( 0,1-\left | y_{i,j}^s-h \right | \right ) \times g_{rad},\\
			  & g_{rad} = \left\{\begin{matrix}
				0,  & \left | x_{i,j}^s-w \right |\geq 1,                              \\
				1,  & w \geq x_{i,j}^s \;\mathrm{and} \left | x_{i,j}^s-w \right |< 1, \\
				-1, & w<x_{i,j}^s \; \mathrm{and} \left | x_{i,j}^s-w \right |< 1.     \\
			\end{matrix}\right.
		\end{aligned}
	\end{equation}
}

Therefore, we can compute the gradient of $x_{i, j}$ with respect to $\theta_{i}$ by $\frac{\partial x_{i,j}^s}{\partial  \theta_i}=-r_j\sin \theta_i$\newc{,	and similarly for} $\frac{\partial v_{i,j}}{\partial  y_{i,j}^s}$ and $\frac{\partial y_{i,j}^s}{\partial  \theta_i}$. Since $f$ is a linear function, the gradient of $\theta_i$ w.r.t. $z_i$ is easily obtained, and the gradient of $\lambda_k$ about $z_i$ is
\newc{
\begin{equation}
	\begin{aligned}
		\frac{\partial z_i}{\partial \lambda_k}=
		\left\{\begin{matrix}
			\frac{\sum_{c=i+1}^{n}\left \lfloor \lambda_{c} \right \rfloor_+}
			{(\sum_{c=0}^{n}\left \lfloor \lambda_{c} \right \rfloor_+)^2}\  & \textrm{if} \ \lambda_{k}>0 \ \textrm{and}\ k\leqslant i, \\
			\frac{-\sum_{c=0}^{i}\left \lfloor \lambda_{c} \right \rfloor_+}
			{(\sum_{c=0}^{n}\left \lfloor \lambda_{c} \right \rfloor_+)^2}\  & \textrm{if} \ \lambda_{k}>0 \ \textrm{and}\ k >  i,       \\
			0\                                                               & \textrm{otherwise.}
		\end{matrix}\right.
	\end{aligned}
\end{equation}
}

Therefore, we can update $\theta_i$ by updating $\lambda_k$. As shown in Figure~\ref{fig:rst} (c), if we fix $\theta_i$ and initialize all $\lambda_k$ equally, the transformed image is sampled equally from different regions of the \nnc{contour} sketch image. By contrast, if $\theta_i$ is learnable, the transformed image will be sampled more from specific regions that include more identity information.

\vspace{0.1cm}

\begin{figure*}[t]
	\centering
	\includegraphics[height=8.5cm]{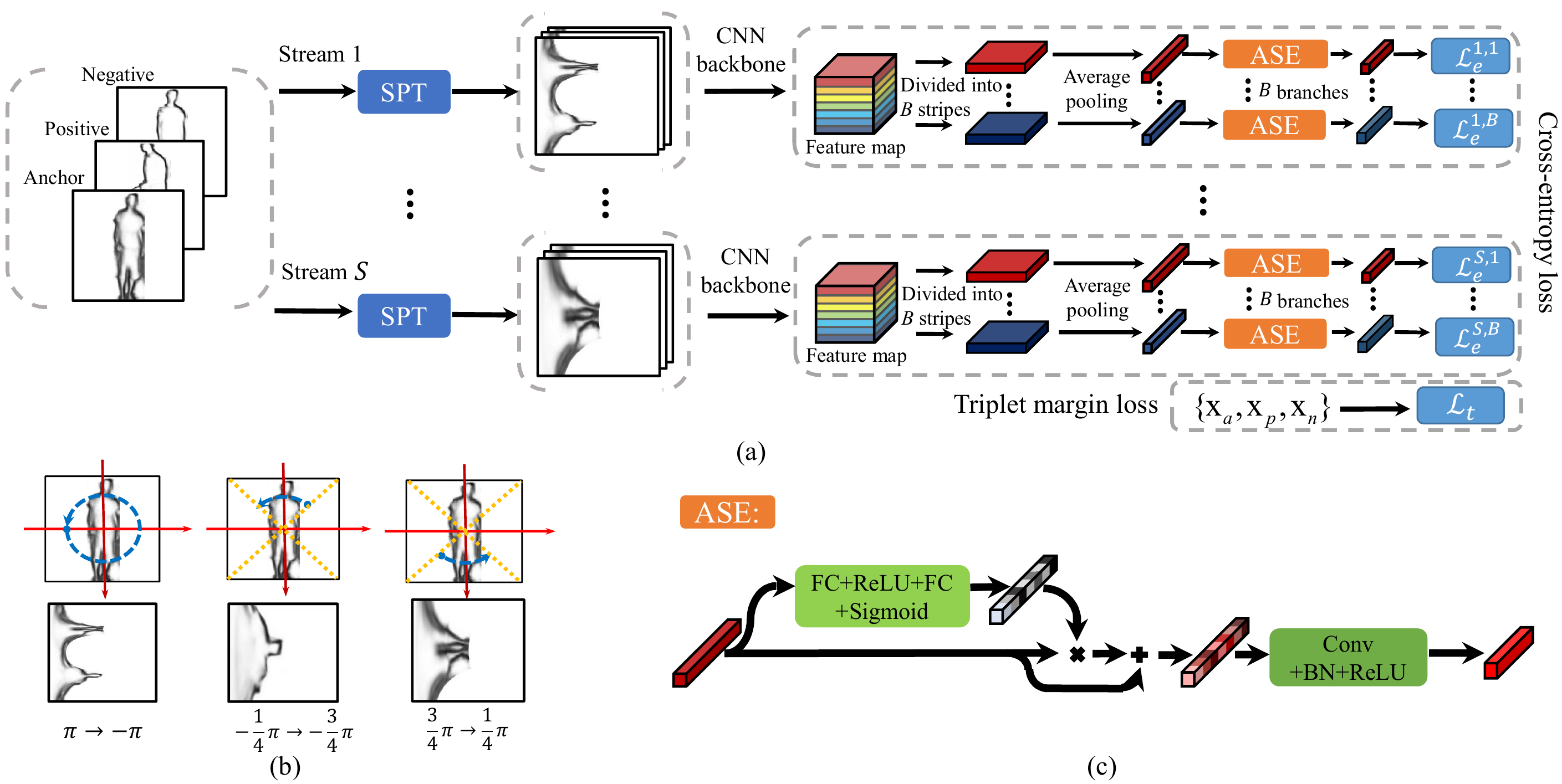}
	\caption{Our multistream model and the details of each component. (a) Our model is a \qzz{multistream} CNN with different SPT layers to transform the \nnc{contour} sketch images into different types. For each stream, \newc{we use the CNN backbone to extract the feature map for each image and divide the feature map into $B$ horizontal angle stripes equally; then we apply the average pooling for each angle stripe and build the unshared ASE components to select discriminative curve patterns by modeling the channel interdependencies.}  The inputs of our model are triplets, and we use cross-entropy loss and triplet margin loss as the loss function. Here, ${\mathbf{x}_a,\mathbf{x}_p,\mathbf{x}_n}$ represent the features of the anchor, positive, and negative images, respectively.
		(b) The \nnc{contour} sketch images are transformed by three different spatial polar transformation (SPT) layers.
		(c) The detailed structure of angle-specific extractor (ASE). 
	}
	\label{fig:flow}
\end{figure*}

\subsubsection{Insight understanding}

This transformation can be viewed as unrolling the \nnc{contour} sketch image with respect to its center, i.e., unrolling the polar coordinate space of the \nnc{contour} sketch image. Each row of the transformed image is sampled from the pixels of the original image with the same polar angle.

\chgorder{
In fact, by using SPT, when the convolutional layer is applied to the transformed image, the receptive field of the convolutional layer is sector-shaped or cone-shaped, instead of rectangular, with respect to the original image,
as shown in Figure~\ref{fig:rst} (d).
}

\qz{\newc{Alternatively,} we also have considered other schemes to select discriminative curve patterns, such as spatial attention \cite{li2018harmonious}, deformable convolution \cite{dai2017deformable} and spatial transformation network \cite{jaderberg2015spatial}. Compared to these methods,
 our SPT selects/transforms the images to the same coordinate space.
We also evaluate these methods in our experiments \newc{(see Section~\ref{sec:result}).}}

\revision{
	Note that although PTN \cite{esteves2018polar} is also based on the polar transformation, it is very different from our SPT. SPT focuses on learning the sampled angles \wss{which facilitates learning multi-granularity features as shown later} and sets the origin of the polar transformation located at the center of human contour sketch image; in comparison, PTN learns the origin of the polar transformation for each image. And thus PTN could learn different transformation origins and therefore obtain different transformed images for different input images, even though they are from the same identity, which is not effective for the re-identification (See Section \ref{section:futher_evaluation}).
}

\subsection{Angle-Specific Extractor}

We wish to further exploit fine-grained \qz{angle}-specific features for robustness against clothing changes across camera views.
Specifically, \nncc{as shown in Figure~\ref{fig:flow} \lastc{(a)},} we split the feature map into \qzz{$B$} angle stripes
based on the corresponding range of angles in the transformed images, followed by an average pooling for each angle stripe.
\qz{Then, the features of different angle stripes are fed into different CNN branches to learn refined and angle-specific features.}

\chgorder{
Since different channels of features in a CNN represent different patterns, that is, different channels contribute differently to recognition, we reweight the channel by modeling the interchannel dependencies. As the interchannel dependencies of different angle strips differ, we use unshared layers to model the dependencies for different angle stripes.
Modeling the interdependencies in such a way is helpful to provide more attentions to relatively invariant curve patterns.
These interchannel dependencies act as channel weights,
which are used to reweight the feature map by the element-wise product
to model the interdependency between channels of the feature map.
}

\newc{Specifically,} \lastc{as shown in Figure~\ref{fig:flow} (c)}, \qzz{for the $j$-th ($j=1,\dots,B$) branch, the interchannel dependency can be computed by a gating mechanism}, including a fully connected layer
with weight $\mathbf{W}_j\in \mathbb{R}^{\frac{L}{r}\times L}$ for dimension reduction, a ReLU activation $\zeta$, a fully connected layer with weight $\mathbf{Q}_j\in \mathbb{R}^{L \times \frac{L}{r}}$ for dimension incrementation and a sigmoid activation $\sigma$. Here, $L$ represents the number of input channels, and $r$ is the reduction ratio. Let $\mathbf{a}_j\in \mathbb{R}^L$ be the input vector \qz{(i.e. the feature of each  angle stripe after average pooling)}; then, the dependencies $\mathbf{d}_j\in \mathbb{R}^L$ \qzz{of the $j$-th branch} can be computed by
\begin{align}
	\mathbf{d}_j=\sigma \left ( \mathbf{Q}_j\zeta \left ( \mathbf{W}_j\mathbf{a}_j \right ) \right ).
\end{align}
\qzz{Since} the channel weights extracted from each angle stripe could be corrupted by local noise.
\newc{Therefore, we introduce a shortcut connection architecture 
with an element-wise sum to reduce the effect of \newc{local} noise.} This process can be formulated as
\begin{align}
	\mathbf{o}_j=\mathbf{a}_j+\mathbf{a}_j\odot\mathbf{d}_j,
\end{align}
\qzz{where $\odot$ represents the element-wise product and $\mathbf{o}_j$ is the output \newc{after reweighting}.} Finally, an additional convolutional layer is applied to adapt the features.

\subsection{Learning \newc{Multi-granularity} Features}



\qzz{To extract the \newc{multi-granularity} features \nncc{for mining global-to-local discriminative feature}, we adopt a multistream CNN  \newc{(as shown in Figure~\ref{fig:flow} (a))} as our
feature extractor by varying the SPT sampling range. \newc{By varying the SPT sampling range, our network is able to exploit the coarse-grained feature for the global image, as well as the local fine-grained feature for the local image.} For this purpose, }
a series of linear functions $f$ in Eq. (\ref{eq:mapping}) is designed to map
$z_i$ to $\theta_i$ in different ranges,
so we can obtain different regions of transformed images.
\qzz{As shown in Figure~\ref{fig:flow} \qzz{(b)}, in this work, we set the stream number $S$ of the multistream network to 3. We obtain
the transformations by setting
$f_1(z_i)=-2\pi \times z_i + \pi$ \qzz{(i.e. $a=\pi, b=-\pi$)} to \newc{form stream 1 for extracting global feature},
$f_2(z_i)=(-\frac{3}{4}\pi + \frac{1}{4}\pi) \times z_i - \frac{1}{4}\pi$ \qzz{(i.e. $a=-\frac{1}{4}\pi, b=-\frac{3}{4}\pi$)} and
$f_3(z_i)=(\frac{1}{4}\pi - \frac{3}{4}\pi) \times z_i + \frac{3}{4}\pi$ \qzz{(i.e. $a=\frac{3}{4}\pi, b=\frac{1}{4}\pi$)} to \newc{form stream 2 and 3 for extracting} local features, where these two regions contain more patterns.
}

\qzz{By learning a series of functions $f$, we can obtain different transformed
images. Then, different streams take different transformed images as input for extracting
features \newc{at multiple granularities.}}

\begin{figure*}[t]
	\centering
	\includegraphics[height=5.2cm,trim=0 0 0 0,clip]{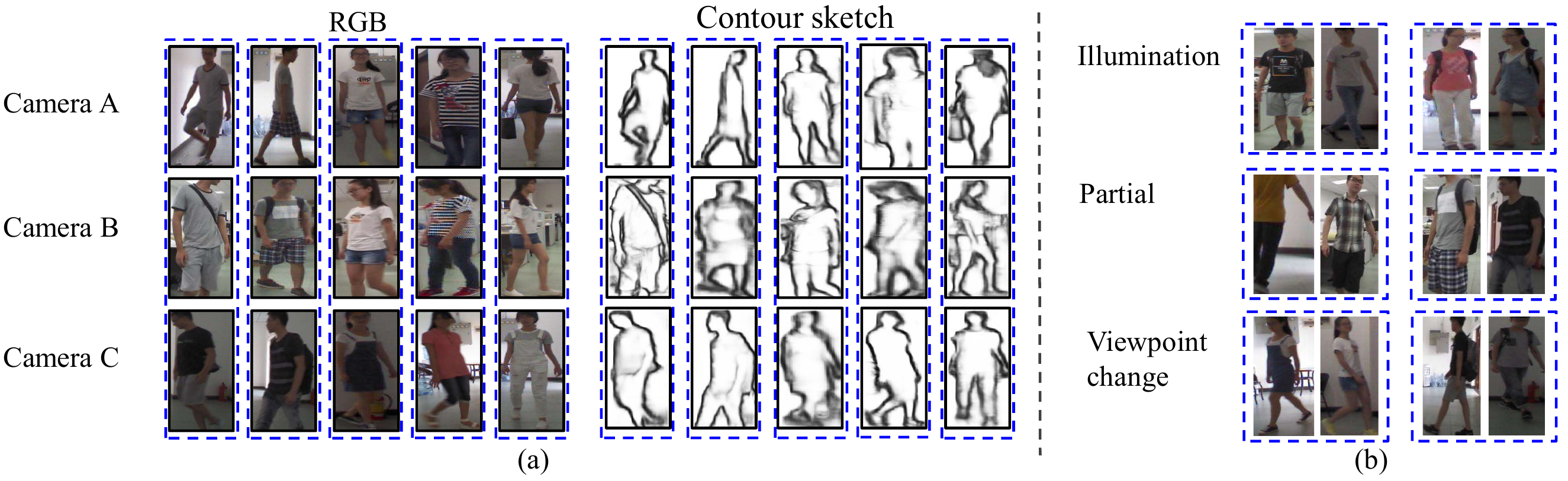}
	\caption{Examples of the PRCC dataset. (a) The left-hand side are RGB images, and the corresponding
		\nnc{contour} sketch images are on the right-hand side, where images of the same column are from the same person. (b) Other variations of our collected dataset. \qzz{The images in the same dash box are of the same identity.}
	}
	\label{fig:dataset}
\end{figure*}

\subsection{Learning Towards Clothing Invariant Features}

While the contour sketch of the same person is relatively consistent under moderate clothing changes, not all contour patterns are robust to the changes. Therefore, we further mine the robust clothing-invariant curve patterns across the contour sketches of the same person under moderate clothing change.
During training, we regard images of the same person with different clothes as having the same identity. Specifically, for each image, let $\mathbf{x}^{i,j}_a$ represent the feature of the $j$-th branch of the $i$-th stream, and let $\mathbf{x}^{i,j}_p$ represent another feature of the same identity with different clothes. Then, $\mathbf{x}^{i,j}_a$ and $\mathbf{x}^{i,j}_p$ are nonlinearly mapped  to a $\mathcal{C}$-dimensional identity prediction distribution by a $\mathrm{softmax}$ function as follows:
\begin{align}
	\mathbf{y}^{i,j} = \mathrm{softmax} (\mathbf{W}^{i,j}\mathbf{x}^{i,j}+\mathbf{b}^{i,j}), \mathbf{x}^{i,j} \in \{\mathbf{x}^{i,j}_a,\mathbf{x}^{i,j}_p\},
\end{align}
\lastc{where $\mathbf{W}^{i,j}$ and $\mathbf{b}^{i,j}$ are the weight and bias of the last fully connected layer, respectively.} The predicted distribution $\mathbf{y}^{i,j}$ is compared with the ground-truth one-hot labels $\mathbf{g}=[g_1,\dots,g_{\mathcal{C}}]$ by the cross-entropy loss, which is defined by
\begin{align}
	\mathcal{L}^{i,j}_e = \sum_{c=1}^{\mathcal{C}}-g_c \log y_c^{i,j} = -\log y_t^{i,j} ,
\end{align}
where $t$ indicates the ground-truth index.

The relation between images of the same person with different clothes is not \qzz{explicitly} considered in the cross-entropy loss, so we \nncc{utilize} the auxiliary triplet margin loss.
Learning with triplets involves training from samples of the form $\left\{\mathbf{x}_a,\mathbf{x}_p,\mathbf{x}_n\right\}$, which represent the concatenated features of all branches and streams of the anchor, positive and negative sample, respectively. Reductions in the intraclass gap and learning \ws{shared latent} feature of the positive pair $\left\{\mathbf{x}_a,\mathbf{x}_p\right\}$ and alleviation of the influence of the negative sample $\mathbf{x}_n$ are beneficial.


Let $d(\mathbf{x},\mathbf{y}) = \left \| \mathbf{x}-\mathbf{y} \right \|_2$; then, the triplet margin loss can be defined as:
\begin{align}
	\mathcal{L}_t =
	\left \lfloor m+d(\mathbf{x}_a,\mathbf{x}_p)-d(\mathbf{x}_a,\mathbf{x}_n) \right \rfloor_+,
\end{align}
where $m$ is the margin of triplet loss.
The final loss is composed by
the cross-entropy loss $\mathcal{L}_e^{i,j}$ and a triplet loss $\mathcal{L}_t$ with a weight $\eta$, which can be written as
\begin{align}
	\mathcal{L} = \sum_{i=1}^{S}\sum_{j=1}^{B}\mathcal{L}_e^{i,j}+\eta \mathcal{L}_t,
\end{align}
where $B$ and $S$ represent the number of branches and the number of streams, respectively.
In our implementation, the margin $m$ and the weight $\eta$ are set to 5.0 and 10.0, respectively.


\subsection{Summary of Our Model and Network Structure}

In summary, as shown in Figure~\ref{fig:flow}, the contour sketch images are first transformed by SPT to focus more attention on relatively invariant and discriminative curve patterns. Our model consists of a series of SPT layers
with different linear functions $f$ to constrain the range of \qqz{$\bm\theta$}; therefore, we
obtain different types of transformed images, which \newc{are} beneficial to learning the
\newc{multi-granularity} features for CNN.
After the SPT layers, we use CNN to extract
the feature maps for each stream and then divide the feature maps into different
horizontal stripes, followed by average pooling. Next, for each stripe,
ASE is applied to extract fine-grained \qz{angle}-specific features.
Note that each stream and each branch have the same structure and do not share parameters.
Finally, we compute the cross-entropy loss for each branch and concatenate the output vector of each branch to compute the triplet loss.

\section{A Cross-clothes Re-id Dataset and Processing}

Existing person re-id datasets, such as VIPER \cite{gray2007evaluating},
CUHK(01,03) \cite{li2012human,li2014deepreid}, SYSU-MM01 \cite{wu2017rgb},
Market-1505 \cite{zheng2015scalable} and \roundtwo{DukeMTMC-reID \cite{ristani2016performance,zheng2017unlabeled}},
are not suitable for testing person re-id under clothing change since the people in these datasets
wear the same clothes in the different camera views.

Only a few pedestrian datasets are publicly available for studying person re-id under clothing change,  and these datasets contain very few identities (e.g., BIWI \cite{munaro2014one} has only 28 persons with clothing change). In this work, we contribute a new person re-id dataset with moderate clothing change, called the Person Re-id under moderate Clothing Change (PRCC) dataset, which is ten times larger than the BIWI dataset in terms of the number of people. The PRCC consists of 221 identities with three camera views. 
As \ws{shown} in Figure~\ref{fig:dataset}, each person in Cameras A and B is wearing the same clothes, but the images are captured in different rooms. For Camera C, the person wears different clothes, and the images are captured \nncc{in} a different day. 
For example, the woman in the fourth column of Camera views A and B is wearing jeans, a striped T-shirt and a pair of red sneakers, whereas in Camera view C, she is wearing a pink T-shirt, shorts and a pair of sandals. Although our proposed method assumes that \nncc{a} person does not change clothes dramatically, we do not constrain the degree of clothing change in our dataset. \revision{The camera network map is shown in Figure~\ref{fig:camera}.}

The images in the PRCC dataset
include not only clothing changes for the same person across different camera views but also other variations,
such as changes in illumination, occlusion, pose and viewpoint.
In general, 50 images exists for each person in each camera view; therefore, approximately 152 images of each person are included in the dataset, for a total of 33698 images.

To generate the \nnc{contour} sketch images,
we use the holistically nested edge detection model proposed in \cite{xie2015holistically} and adopt the fused
output as our \nnc{contour} sketch images, as shown in
Figure~\ref{fig:dataset}.

In our experiments, we randomly split the dataset into a training set and a testing set. The training set consist of 150 people, and the testing set consist of 71 people, with no overlap between the training and testing sets in terms of identities. During training, we selected
25 percent of the images from the training set as the validation set. Our dataset is available at: \url{http://www.isee-ai.cn/%7Eyangqize/clothing.html}





\begin{figure}
	\centering
	\includegraphics[height=5cm,trim=0 0 0 0,clip]{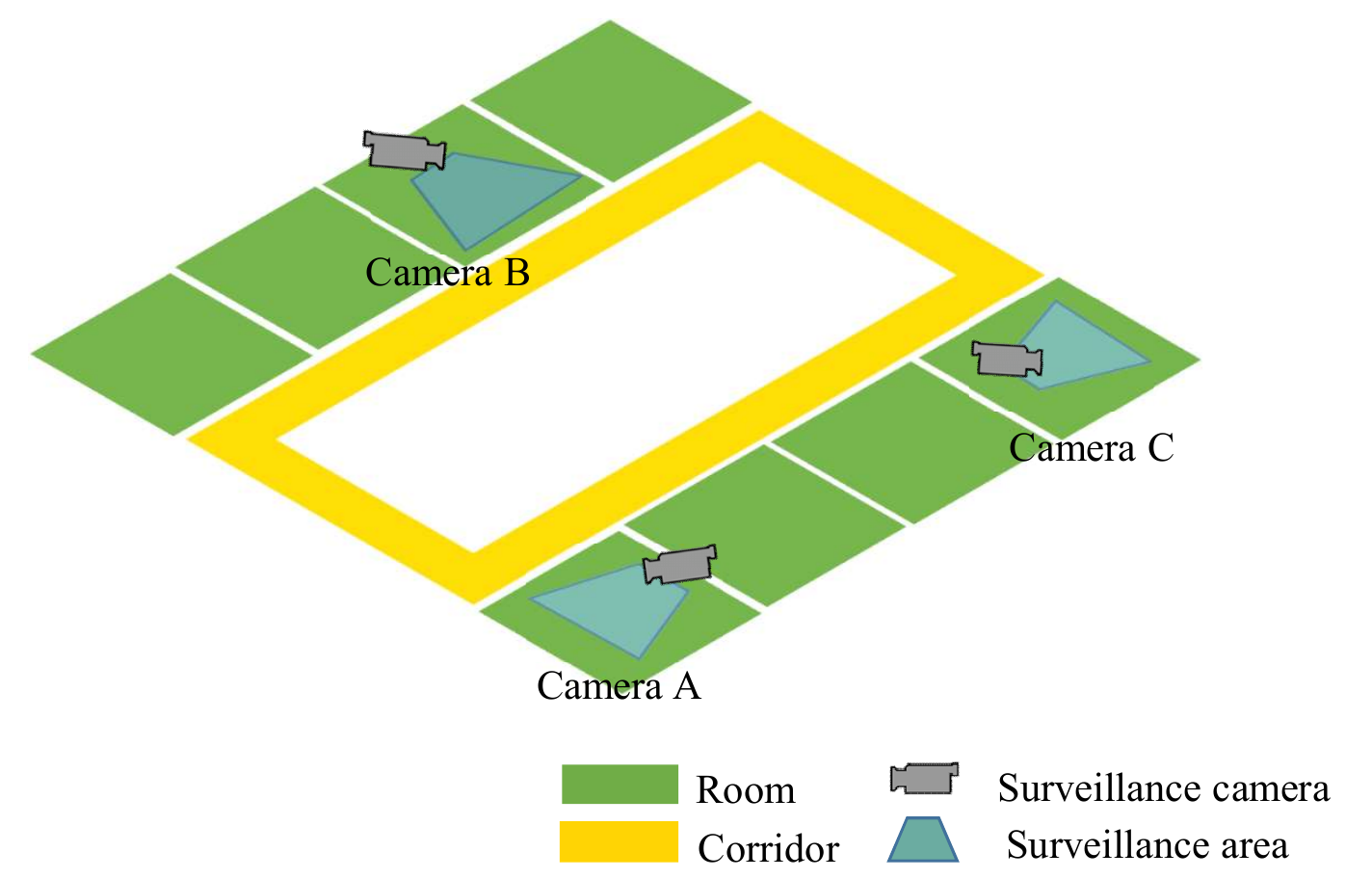}
	\caption{\revision{A diagrammatic layout of the camera network map of the PRCC dataset.}}
	\label{fig:camera}
\end{figure}

\section{Experiments}

\subsection{Implementation Details}

\noindent\textbf{Network Details.} 
\revision{Each contour sketch image is padded to a square of which the side is the same as the height of person image (i.e. the largest side of person image) with the empty filled by a pixel value of 255, so it does not change the shape of the body (i.e. the ratio between the height and width of a person body does not change). Then these padded images are resized to a resolution of $224 \times 224$.}
The maximum radius $R$ is 2, and the original point is set in the center of the \nnc{contour} sketch image. We set the initial value of all $\lambda_k$ to 0.05. We use Res-Net18 \cite{he2016deep} as our backbone network, but we remove the last \qzz{residual block} so that the output channel $C$ is equal to 256. \qzz{We divide the feature map into 7 horizontal stripes equally so that the number of CNN branches of each stream is 7 for each stream}, and the reduction ratio $r$ is 2. Finally, we compute the cross entropy-loss for each stream. During testing, we concatenate the output of each stream as the final feature of the contour sketch image.

\noindent\textbf{Training.} During training, we set the batch size to $64\times3$.
First, we fixed $\theta_i$ and trained the whole network for 30 epochs and then jointly trained each
SPT layer and the network for 60 epochs. Next, we fixed the sampled angle $\theta_i$ of the SPT
layer and optimized only the parameters of each stream network for 30 epochs. The learning
rate of the network was set to 0.1 initially, decaying 0.1 every 30 epochs. The
learning rate of the SPT was set to 0.0001 initially, decaying 0.1 every 30 epochs.
We used SGD \cite{sutskever2013importance} with momentum as our optimizing algorithm, with the weight decay and momentum set to $10^{-4}$ and $0.9$, respectively.
We concatenated the output of each branch of each stream as the final feature to compute the triplet margin loss.

\vspace{0.1cm}

\noindent\textbf{Testing and Evaluation.} The testing set was randomly divided into a gallery set and a probe set. For every identity in the testing set, we randomly chose one image in Camera view A to form the gallery set for single-shot matching. All images in Camera views B and
camera C were used for the probe set. Specifically, the person matching between Camera views A and B was performed without clothing
changes, whereas the matching between Camera views A and C was cross-clothes matching.
The results were assessed in terms of the cumulated matching characteristics, specifically, the rank-\textit{k}  matching accuracy.
We repeated the above evaluation 10 times with a random selection of the gallery set and computed the
average performance.

\begin{table*}[t]
	\setlength{\tabcolsep}{16pt}
	\begin{center}
		\caption{Performance (\%) of our approach and the compared methods on the PRCC dataset. \qz{``RGB'' means the inputs of the model are RGB images; ``Sketch''  means the inputs of the model are contour sketch images}}
		\label{table:result}
		\resizebox{0.93\textwidth}{!}{%
		\begin{tabular}{l|c|c|c||c|c|c}
			\hline
			\multirow{2}{*}{Methods}
			&\multicolumn{3}{c||}{\tabincell{c}{Cameras A and C (Cross-clothes)}}
			&\multicolumn{3}{c}{\tabincell{c}{Cameras A and B (Same clothes)}} \\ \cline{2-7}
			                                                                                                     & Rank 1         & Rank 10 & Rank 20 & Rank 1         & Rank 10 & Rank 20 \\\hline
			LBP \cite{ojala1996comparative} + KISSME \cite{koestinger2012large}                                  & 18.71          & 58.09   & 71.40   & 39.03          & 76.18   & 86.91   \\
			HOG \cite{dalal2005histograms} + KISSME \cite{koestinger2012large}                                   & 17.52          & 49.52   & 63.55   & 36.02          & 68.83   & 80.49   \\
			LBP \cite{ojala1996comparative} + HOG \cite{dalal2005histograms} + KISSME \cite{koestinger2012large} & 17.66          & 54.07   & 67.85   & 47.73          & 81.88   & 90.54   \\

			LOMO \cite{liao2015person} + KISSME \cite{koestinger2012large}                                       & 18.55          & 49.81   & 67.27   & 47.40          & 81.42   & 90.38   \\\hline
			LBP \cite{ojala1996comparative} + XQDA \cite{liao2015person}                                         & 18.25          & 52.75   & 61.98   & 40.66          & 77.74   & 87.44   \\
			HOG \cite{dalal2005histograms} + XQDA \cite{liao2015person}                                          & 22.11          & 57.33   & 69.93   & 42.32          & 75.63   & 85.38   \\
			LBP \cite{ojala1996comparative} + HOG \cite{dalal2005histograms} + XQDA \cite{liao2015person}        & 23.71          & 62.04   & 74.49   & 54.16          & 84.11   & 91.21   \\

			LOMO \cite{liao2015person} + XQDA \cite{liao2015person}                                              & 14.53          & 43.63   & 60.34   & 29.41          & 67.24   & 80.52   \\\hline
			Shape Context \cite{belongie2002shape}                                                               & 11.48          & 38.66   & 53.21   & 23.87          & 68.41   & 76.32   \\
			LNSCT \cite{xie2010extraction}                                                                       & 15.33          & 53.87   & 67.12   & 35.54          & 69.56   & 82.37   \\\hline
			Alexnet \cite{krizhevsky2012imagenet} (RGB)                                                          & 16.33          & 48.01   & 65.87   & 63.28          & 91.70   & 94.73   \\
			VGG16 \cite{simonyan2014very} (RGB)                                                                  & 18.21          & 46.13   & 60.76   & 71.39          & 95.89   & 98.68   \\
			Res-Net50 \cite{he2016deep} (RGB)                                                                    & 19.43          & 52.38   & 66.43   & 74.80          & 97.28   & 98.85   \\
			HA-CNN \cite{li2018harmonious} (RGB)                                                                 & 21.81          & 59.47   & 67.45   & 82.45          & 98.12   & 99.04   \\
			PCB \cite{sun2017beyond} (RGB)                                                                       & 22.86          & 61.24   & 78.27   & \textbf{86.88} & 98.79   & 99.62   \\\hline
			Alexnet \cite{krizhevsky2012imagenet} (Sketch)                                                       & 14.94          & 57.68   & 75.40   & 38.00          & 82.15   & 91.91   \\
			VGG16 \cite{simonyan2014very} (Sketch)                                                               & 18.79          & 66.01   & 81.27   & 54.00          & 91.33   & 96.73   \\
			Res-Net50 \cite{he2016deep} (Sketch)                                                                 & 18.39          & 58.32   & 74.19   & 37.25          & 82.73   & 93.08   \\
			HA-CNN \cite{li2018harmonious} (Sketch)                                                              & 20.45          & 63.87   & 79.58   & 58.63          & 90.45   & 95.78   \\
			PCB \cite{sun2017beyond} (Sketch)                                                                    & 22.48          & 61.07   & 77.05   & 57.36          & 92.12   & 96.72   \\\hline
			SketchNet  \cite{zhang2016sketchnet} (Sketch+RGB)                                                    & 17.89          & 43.70   & 58.62   & 64.56          & 95.09   & 97.84   \\\hline
			Face  \lastcc{\cite{wen2016discriminative}}                                   & 2.97           & 9.85    & 13.52   & 4.75           & 13.40   & 45.54   \\\hline
			Deformable Conv. \cite{dai2017deformable}                                                            & 25.98          & 71.67   & 85.31   & 61.87          & 92.13   & 97.65   \\
			STN \cite{jaderberg2015spatial}                                                                      & 27.47          & 69.53   & 83.22   & 59.21          & 91.43   & 96.11   \\\hline
			Our model                                                                                            & \textbf{34.38} & 77.30   & 88.05   & \textbf{64.20} & 92.62   & 96.65   \\
			\hline
		\end{tabular}
		}
	\end{center}
\end{table*}

\subsection{Results on the PRCC Dataset}
\label{sec:result}

\noindent\textbf{Compared methods}.
We evaluated three \qqz{hand-crafted} features that are typically used for representing texture information, namely,
HOG \cite{dalal2005histograms}, LBP \cite{ojala1996comparative} and LOMO \cite{liao2015person}.
These \qqz{hand-crafted} feature are enhanced by metric learning models, such as
KISSME \cite{koestinger2012large} and XQDA \cite{liao2015person}.
\qqz{We compared with LNSCT \cite{xie2010extraction} to evaluate whether the \newc{contourlet-based} feature is effective for clothing changes}. Since our contour-sketch-based method is related to shape matching, we also compared ``Shape Context'' \cite{belongie2002shape}.

Since CNN has achieved considerable success in image classification, we also tested several common
CNN structures, including Alexnet \cite{krizhevsky2012imagenet},
VGG16 \cite{simonyan2014very} and Res-Net50 \cite{he2016deep}.
\qqz{We also evaluated a multi-level-attention model HA-CNN \cite{li2018harmonious} and a recent strong CNN baseline model PCB \cite{sun2017beyond}}.
The above deep methods were evaluated on both \nnc{contour} sketch images and RGB images.

We also tested the sketch retrieval method SketchNet \cite{zhang2016sketchnet}.
Since SketchNet was designed for cross-modality search, which requires pairwise samples of sketch and RGB images,
we paired a \nnc{contour} sketch image and the corresponding RGB image as a positive pair and randomly paired a \nnc{contour} sketch image with an
RGB image from another identity as a negative pair. 
\nncc{For SketchNet, the RGB images and the contour sketch images are resized to the same size to maintain the spatial structure.}

Our SPT is related to the convolution strategy,
we also \ws{compared} deformable convolution \cite{dai2017deformable} to demonstrate the effectiveness of \ws{SPT in} our proposed cross-clothes person re-id method.
Deformable convolution can change the convolution kernel shape to focus on the
curves and achieve better performance.
We report the performance of deformable convolution by removing
our SPT layers and using the deformable convolutional layer to replace the second
convolutional layer of our model.
Since our multistream model \newc{selects discriminative curve patterns and} extracts \newc{multi-granularity} features,
we also use STN \cite{jaderberg2015spatial} to replace the SPT layer in our network to validate the effectiveness of our proposed transformation. In this case, STN serves as an affine transformation.

\nncc{Finally, as the face cues are independent of clothing changes, 
\lastcc{we compared a standard face recognition method \cite{wen2016discriminative}, which has an accuracy of 99.28\% on the Labeled Faces in the Wild benchmark \cite{huang2008labeled}}}.

\vspace{0.1cm}

\noindent\textbf{Results and Discussion}.
The experimental results are reported in Table~\ref{table:result}. Our
proposed method achieves the best rank-1 accuracy (\textbf{34.38\%}) among the compared methods,
including \qqz{hand-crafted} features, deep-learning-based methods and the strong baseline model PCB \cite{sun2017beyond}, for person
re-id under moderate clothing change.
\chgorder{
The performance of PCB (RGB) on our dataset indicates that the problem we consider in this paper is very challenging for person re-id.
PCB achieved rank-1 accuracy of 92.3\% \lastc{\cite{sun2017beyond}} on Market-1501 that has no clothing change between person images from the same identity but only
22.86\% on our dataset in the cross-clothes matching scenario.}


\newc{LNSCT \cite{xie2010extraction} and ``Shape Context'' \cite{belongie2002shape} are the representative handcrafted contour features. However, these methods are not designed for cross-clothing person re-id and lack modeling the view and clothing changes.}
Our proposed model also performed better than \qz{HA-CNN, STN and deformable convolution},
and the performance of CNN on original \nnc{contour} sketch images was unsatisfactory.
These results suggest that our proposed SPT transformation is effective for
helping to extract reliable and discriminative features from a \nnc{contour} sketch image.


Although person re-id with no clothing change (i.e. ``Same Clothes'' in the Table \ref{table:result})
is not the aim in this work, when the input images
are \nnc{contour} sketch images, our method can still achieve an accuracy of \textbf{64.20\%}, which
is better than that of hand-crafted features with metric learning methods \qqz{and deep learning methods (including
	Alexnet \cite{krizhevsky2012imagenet}, VGG16 \cite{simonyan2014very}, Res-Net50 \cite{he2016deep}, HA-CNN \cite{li2018harmonious} and PCB \cite{sun2017beyond}) when the input images are contour sketch images.} When the
input images \nnc{of these deep learning methods} are RGB images, the performance of our method ranked fifth.

\begin{figure*}
	\centering
	\includegraphics[height=3.7cm,trim=0 0 0 0,clip]{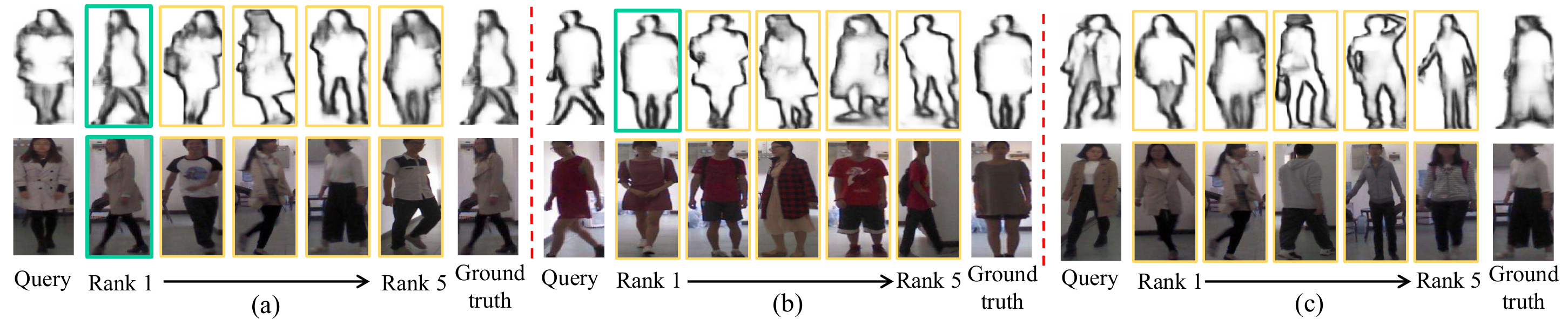}
	\caption{The visualization of the ranking lists of our
	proposed model (top) and PCB (RGB) (bottom). The green box indicates the correct matching. In each subfigure, the
		leftmost image is the probe image, and the rightest is the \newc{ground truth} image \newc{in the gallery set} with the same ID \newc{of the probe image}.
		The middle 5 images are the rank-1 to rank-5 matching images, from left to right.}
	\label{fig:visualization}
\end{figure*}

\begin{table*}[!t]
	\begin{center}
		\setlength{\tabcolsep}{10pt}
		\caption{The statistics on the test set of the PRCC dataset and the corresponding testing result (\%). \ws{\qqz{``Others'' means ``the changes of bags, shoes, hats, haircut''}}
		}
		\label{table:degree}
		\resizebox{\textwidth}{!}{%
		\begin{tabular}{c|c|c||c||c|c|c||c|c|c}
			\hline
			\multicolumn{3}{c||}{\qzz{Type of clothing change}}
				&\multirow{2}{*}{\qzz{\tabincell{c}{The number\\ of identities}}}
				&\multicolumn{3}{c||}{PCB \cite{sun2017beyond} (RGB)}
				&\multicolumn{3}{c}{Our model} \\ \cline{1-3}\cline{5-10}
			\tabincell{c}{\qzz{Upper body}}        
				&\tabincell{c}{\qzz{Lower body}}          
				&\qqz{Others}          
				&
			    & Rank 1 & Rank 10 & Rank 20 & Rank 1 & Rank 10 & Rank 20 \\\hline
			$\surd$ & $\times$ & $\times$ & 23 & 24.47  & 78.65   & 89.83   & 36.04  & 81.96   & 92.67   \\
			$\surd$ & $\surd$  & $\times$ & 24 & 20.14  & 56.63   & 76.05   & 32.67  & 77.60   & 89.80   \\
			$\surd$ & $\surd$  & $\surd$  & 12 & 9.58   & 36.39   & 56.66   & 19.28  & 58.03   & 72.62   \\
			\hline
		\end{tabular}
		}
	\end{center}

\end{table*}

\nncc{From the above experiments, we find} when the input images are RGB images without clothing changes, VGG16 \cite{simonyan2014very}, Res-Net50 \cite{he2016deep}, \qqz{HA-CNN \cite{li2018harmonious}} and
PCB \cite{sun2017beyond} achieve good performance, but they have a sharp performance drop when a clothing change occurs, illustrating the challenge of person re-id when a person dresses differently. The application of existing person re-id methods is not straightforward in this scenario. \qz{Nevertheless, these methods still suggest that the attention mechanism (e.g. HA-CNN) and the fine-grained feature learning (e.g. PCB) are beneficial to learn clothing invariant feature.}

\nncc{Finally, we conducted an interesting investigation by comparing our model with a standard face recognition \lastcc{\cite{wen2016discriminative}} for overcoming the clothing change problem.}
\newc{For comparison with face recognition, given a set of gallery person images and a query image, we first detected the face \lastcc{using \cite{amos2016openface}}, and then extracted the face feature and computed the pairwise distance to get the ranking list.
\lastcc{If the face detection in query image or gallery image fails, the matching associated to that query image or gallery image will be considered as a failure.}
}
The face recognition method performed worst on our dataset. The challenges for applying face recognition methods to surveillance include the low resolution of person images, occlusion, illumination, and viewpoint (e.g. back/front) changes, \lastcc{which could incur the loss of face observation}. These challenges make it difficult for the compared face recognition method to detect faces and extract discriminative features \newc{(the average resolution of the detected faces is about 30/25 pixels in height/width) on our person re-id dataset, which was captured at a distance}.
\chgorder{Our experimental results also validated the challenge of face detection and recognition in the real-world surveillance applications, and currently, the best rank-1 accuracy of surveillance face recognition is \nnc{29.9\%} as \nncc{recently} reported on the QMUL-SurvFace benchmark \cite{cheng2018surveillance}}.

\begin{table}[!t]
	\begin{center}
		\caption{The performance (\%) of PCB (RGB) on the subset of the PRCC dataset {by using only either the upper body information or the lower one when only the clothing changes happen only on the upper body} (23 identities as mentioned in Table~\ref{table:degree}.)
		}
		\label{table:rgb}
		\begin{tabular}[t]{|l||c|c|c|}
			\hline
			\multirow{2}{*}{Part of body}&\multicolumn{3}{c|}{PCB \cite{sun2017beyond} (RGB)} \\ \cline{2-4}
			                            & Rank 1 & Rank 10 & Rank 20 \\\hline
			\qqz{Only using upper body} & 2.02   & 19.15   & 27.62   \\\hline
			\qqz{Only using lower body} & 27.24  & 74.89   & 89.40   \\

			\hline
		\end{tabular}
	\end{center}
\end{table}

\subsection{Analysis of the Degree of Clothing Change}
To further analyze clothing changes and our assumption,
we divide the clothing changes in our dataset into three types, i.e., upper-body clothes,
lower-body clothes and \qqz{others} (including bags, shoes, hats, haircut).
In this way, we can measure the extent of the clothing change and
the corresponding impact on performance, which is shown in Table~\ref{table:degree}. \nncc{Since PCB
(RGB) achieved the best performance among the compared methods
that take RGB images as input, we only compared with PCB
(RGB) here and in the following experiments.}

In our experiments, we find that the proposed methods are more stable than the compared methods. 
If \nncc{people} only \nncc{change} their upper-body clothes, the RGB-based methods still perform well since the features of the lower-body clothes are \ws{still useful to identify the person; and this is validated by an additional experiment as shown in Table~\ref{table:rgb}.}
However, if the person completely changes clothes, the RGB-based methods are not effective. \ws{Note that the RGB-based methods do not fail entirely in this case, because the RGB images also contain some contour information.}

\begin{figure*}[t]
	\centering
	\includegraphics[height=7.3cm,trim=0 0 0 0,clip]{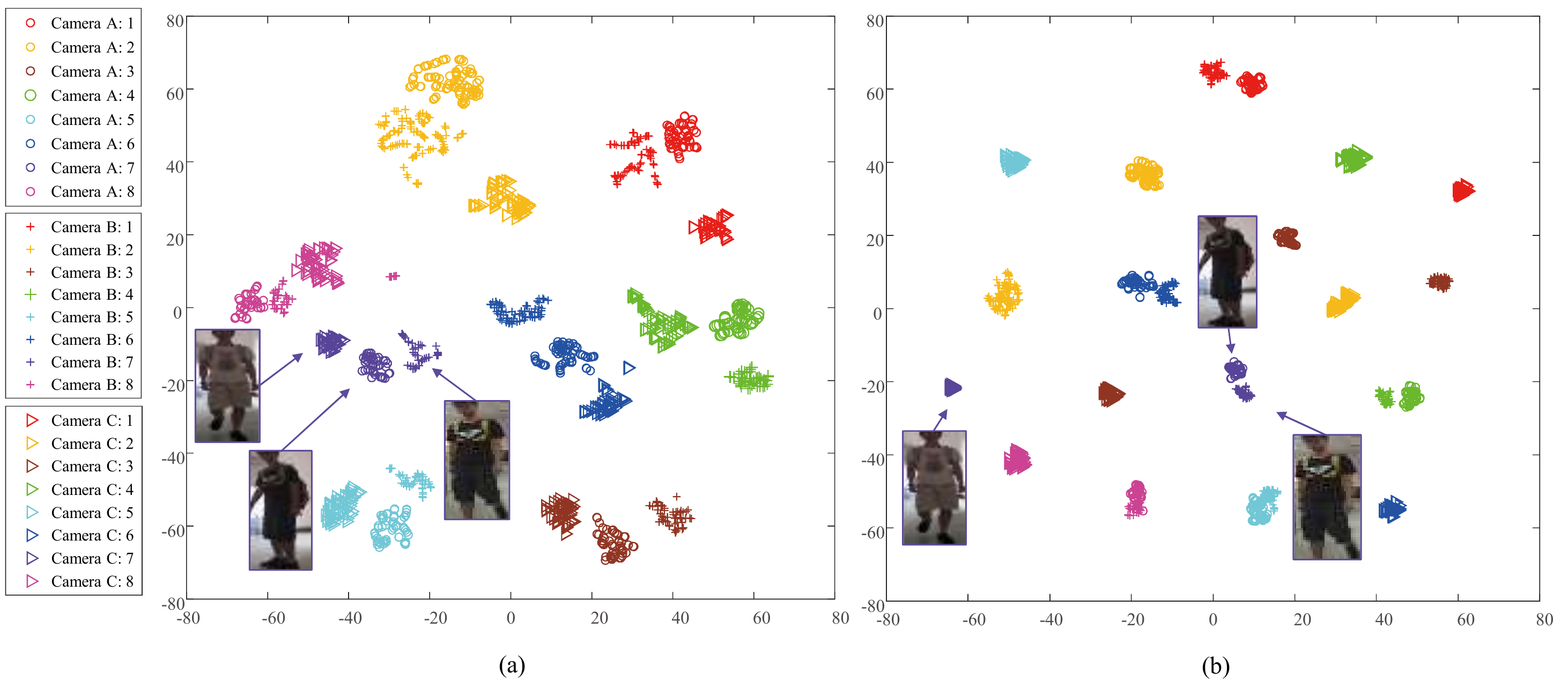}
	\caption{\revision{The t-SNE visualizations for the features of our proposed method (i.e. (a)) and Res-Net50 (i.e. (b)) on the PRCC dataset. Each color represents an identity which is randomly chosen from the testing set. Each symbol (circle, triangle, or cross) represents the camera label indicating where an image is captured. The person is wearing the same clothing in Camera A and B, and he/she wears different clothes in Camera C. The numbers in the legend are used to indicate the identities.}}
	\label{fig:tsne_visualization}
\end{figure*}

We also visualize the ranking lists of the results
to validate the aforementioned assumption in Subsection 3.1. Figure~\ref{fig:visualization} (a-b) shows that our contour-sketch-based method can identify the
target person even when the person changes clothes, whereas the RGB-based methods are inclined to match other persons who are dressed in similar clothes.
An example of a failure of our method is shown in Figure~\ref{fig:visualization} (c).
Because the person changed her clothes dramatically (i.e., changed
from leggings to slacks, added a coat, and changed shoes),
our method failed.
\nncc{Note that our design is for moderate clothing change;} thus, our method is \nncc{still} limited if a person changes his/her clothes substantially,
leading to large variation in the shape of the person.

\subsection{\revision{Visualization of RGB Feature and Contour Sketch Feature}}
\revision{
	We use t-SNE \cite{maaten2008visualizing} to visualize the final learned features for our proposed method (using contour sketch image as input) and Res-Net50 (using the color image as input) on the PRCC dataset in Figure \ref{fig:tsne_visualization}. We can see that the features of different clothes of the same identity stay closely in Figure \ref{fig:tsne_visualization} (a), but most are not for color appearance feature as shown in Figure \ref{fig:tsne_visualization} (b) (e.g. persons ``1'', ``4'', ``5'', ``7'', ``8'' in Figure \ref{fig:tsne_visualization} (b)). The color appearance-based methods will be misled by color information; and therefore, as shown in Figure \ref{fig:tsne_visualization} (b), images of the same person who wears different clothes distribute farther away from each other, as compared to the case of a person wearing the same clothes.	
}

\subsection{Evaluation with Large Viewpoint Changes}

We further validate our method under the combination of \nnc{large} viewpoint variation and clothing change on the PRCC dataset. For the same person, we manually removed the side view images in the gallery set such that  only front and back view images are included in the gallery set. Conversely, we removed the front or back view images in the query set, retaining only the side view images. The results are shown in Table~\ref{table:viewpoint}. Although the performance of our proposed method degraded to some extent due to dramatic viewpoint changes, it substantially outperformed the best RGB-based method. In the case of clothing change, the contour is relatively stable compared to color-based visual cues because clothing changes are unpredictable and hard to model, whereas the contour change is relatively \ws{more} predictable.

\begin{table}[t]
	\setlength{\tabcolsep}{12pt}
	\begin{center}
		\caption{The performance (\%) of cross-clothes matching under large viewpoint changes.}
		\label{table:viewpoint}
		\resizebox{\textwidth}{!}{%
		\begin{tabular}[t]{|l||c|c|c|}

			\hline
			\multirow{2}{*}{Methods}
			&\multicolumn{3}{c|}{\tabincell{c}{Camera A and C\\(Cross-viewpoint evaluation)}} \\ \cline{2-4}

			                               & Rank 1 & Rank 10 & Rank 20 \\\hline
			PCB \cite{sun2017beyond} (RGB) & 16.84  & 54.96   & 71.94   \\\hline
			Our model                      & 25.33  & 67.93   & 83.09   \\

			\hline
		\end{tabular}
		}
	\end{center}
\end{table}

\begin{table*}[t]
	\begin{center}
		\setlength{\tabcolsep}{12pt}
		\caption{The results (\%) of person re-id with clothing change under occlusion on the PRCC dataset. \qz{``0.5 $\Rightarrow$ 0.6'' means the images of probe set are cropped  to 50\% size of the original \qzz{and} the images of the gallery set are cropped to 60\% size of the original. Others are similar.}}
		\label{table:occlusion}
		\resizebox{\textwidth}{!}{%
		\begin{tabular}{c||c|c|c|c||c|c|c|c}
			\hline
			\multirow{2}{*}{Methods}
			&\multicolumn{4}{c||}{\lastc{Between cropped images}}
			& \multicolumn{4}{c}{\lastc{Between cropped and uncropped images}} \\ \cline{2-9}
			  & Matching              & Rank 1 & Rank 10 & Rank 20 & Matching            & Rank 1 & Rank 10 & Rank 20 \\ \hline
			\multirow{5}{*}{PCB \cite{sun2017beyond} (RGB)}&
			0.5 $\Rightarrow$ 0.5 & 11.53  & 41.32   & 59.33   & 0.5 $\Rightarrow$ 1 & 13.42  & 47.49   & 65.91   \\
			  & 0.6 $\Rightarrow$ 0.6 & 13.47  & 46.01   & 63.28   & 0.6 $\Rightarrow$ 1 & 15.45  & 49.84   & 67.67   \\
			  & 0.7 $\Rightarrow$ 0.7 & 15.25  & 49.41   & 66.40   & 0.7 $\Rightarrow$ 1 & 16.58  & 51.10   & 68.82   \\
			  & 0.8 $\Rightarrow$ 0.8 & 16.63  & 51.80   & 68.75   & 0.8 $\Rightarrow$ 1 & 17.65  & 52.24   & 70.27   \\
			  & 0.9 $\Rightarrow$ 0.9 & 18.29  & 52.81   & 70.91   & 0.9 $\Rightarrow$ 1 & 18.78  & 53.41   & 71.24   \\

			\hline
			\multirow{5}{*}{Our model}&
			0.5 $\Rightarrow$ 0.5 &12.06   &43.73    &60.65    & 0.5 $\Rightarrow$ 1 &18.06  &56.87  &71.77  \\
			  & 0.6 $\Rightarrow$ 0.6 & 16.88  & 52.83   & 68.00   & 0.6 $\Rightarrow$ 1 & 21.15  & 59.17   & 74.36   \\
			  & 0.7 $\Rightarrow$ 0.7 & 19.94  & 57.05   & 72.85   & 0.7 $\Rightarrow$ 1 & 22.66  & 61.49   & 76.53   \\
			  & 0.8 $\Rightarrow$ 0.8 & 22.81  & 60.82   & 75.59   & 0.8 $\Rightarrow$ 1 & 24.53  & 62.76   & 77.52   \\
			  & 0.9 $\Rightarrow$ 0.9 & 25.31  & 63.92   & 78.20   & 0.9 $\Rightarrow$ 1 & 26.09  & 64.21   & 78.28   \\
			\hline
		\end{tabular}
		}
	\end{center}
\end{table*}

\begin{table}[t]
	\begin{center}
		\caption{The rank-1 accuracy (\%) of our approach in the ablation study
			Note that when removing the SPT layers, the model only remains one stream.}
		\label{table:ablation}
		\resizebox{\textwidth}{!}{%
			\begin{tabular}{|l||c|c|c|}
				\hline
				\multirow{2}{*}{Methods}
				&\multicolumn{3}{c|}{\tabincell{c}{Camera A and C\\ (Cross-clothes)}} \\ \cline{2-4}
												 & Rank 1         & Rank 10 & Rank 20 \\\hline
				Fixed $\bm{\theta}$ of SPT       & 31.05          & 72.68   & 86.79   \\
				Removing SPT                     & 25.74          & 70.66   & 83.08   \\\hline
				Removing ASE                     & 28.00          & 70.89   & 84.11   \\\hline
				Removing $\mathcal{L}_t$         & 31.39          & 72.62   & 85.00   \\\hline	
				Removing ASE, SPT, $\mathcal{L}_t$ & 21.95          & 56.81   & 73.77   \\
				\hline
				Our full model                   & \textbf{34.38} & 77.30   & 88.05   \\
				\hline
			\end{tabular}
		}
	\end{center}
\end{table}

\subsection{\revision{Evaluation with Partial Body}}
Occlusion results in a camera capturing only a partial observation of the target person, which is generally addressed as partial re-id \cite{zheng2015partial}. We randomly cropped the images to sizes from 50\% to 100\% of the original image \newc{to increase the number of partial samples for} train our model. \revision{Specifically, we first obtain the random height and width of the cropped image (i.e. the ratio of height and width is not fixed). Then, we can crop a region from the original image by randomly selecting the cropping location. Finally}, we extracted images of the testing set by randomly cropping each image into a size from 50\% to 100\% of the original. In this evaluation, we not only performed the matching between cropped images, but also the matching between cropped and uncropped images.
Table~\ref{table:occlusion} shows that as the cropping increases, the performance of our model decreases due to the loss of information.
Additionally, occlusion leads to misalignment matching, e.g., upper body to lower body. \nncc{However, our methods still outperformed the RGB-based methods, even with occlusion, because the RGB-based methods face additional problems (i.e. color information changing). For example, if we only captured the upper body of a person and this person coincidently changed the upper clothes; then this person cannot be identified by color information in this situation because the color information changes thoroughly.} A more promising way to solve the occlusion problem is the combination of part alignment and local-to-local or global-to-local matching \cite{zheng2015partial}. However, solving the occlusion problem is beyond the scope of this work.



\subsection{Ablation Study of the Proposed Model}

\begin{table}[t]
	\begin{center}
		\caption{The performance (\%) of each stream \lastc{of our model}.}
		\label{table:stream}
		\resizebox{\textwidth}{!}{%
			\begin{tabular}{|l||c|c|c|}

				\hline
				\multirow{2}{*}{Combination}
				&\multicolumn{3}{c|}{\tabincell{c}{Camera A and C\\ (Cross-clothes)}} \\ \cline{2-4}

				                            & Rank 1         & Rank 10 & Rank 20 \\\hline
				Stream 1 \newc{(from $\pi$ to $-\pi$)}                & 28.71          & 72.69   & 86.02   \\
				Stream 2 \newc{(from $-1/4\pi$ to $-3/4\pi$)}         & 18.83          & 63.30   & 80.46   \\
				Stream 3 \newc{(from $3/4\pi$ to $1/4\pi$)}                   & 19.17          & 55.93   & 72.11   \\\hline
				All streams (\nncc{i.e.} our full model) & \textbf{34.38} & 77.30   & 88.05   \\
				\hline
			\end{tabular}
		}
	\end{center}
\end{table}

\vspace{0.1cm}

\noindent \textbf{Effect of SPT.} \qzz{As shown in Table~\ref{table:ablation}, the rank-1 accuracy would degrade to 31.05\% or 25.74\% in the cross-clothes matching scenario if we fixed the $\bm{\theta}$ during training or removed the SPT, \newc{respectively}. These experiments validated the effectiveness of our proposed transformation.} Additionally, by designing different $f(z)$ for the SPT layers, we can obtain different types of transformed images, so the model can learn \newc{multi-granularity} features. If we remove these SPT layers, we have only the original \nnc{contour} sketch image as our input, and our network learns only one \newc{granularity} of feature.

\vspace{0.1cm}

\noindent \textbf{Effect of ASE.}
The results shown in Table~\ref{table:ablation} indicate that the introduced angle-specific extractor can improve the performance, since our angle-specific extractor is customized to extract more \newc{angle} specific and distinguishing feature for each stripe.

\vspace{0.1cm}

\noindent \textbf{Effect of triplet loss.}
\qz{As shown in Table~\ref{table:ablation}, if without the triplet loss, the performance of our model would drop to 31.39\% in the cross-clothes matching scenario.} Although the triplet loss is adopted as an auxiliary loss, we can observe the effectiveness of learning shared latent clothing invariant feature.

\vspace{0.1cm}

\noindent \textbf{Removing SPT, ASE and triplet loss (i.e., convolving \nnc{contour} sketch images directly).} After removing SPT, ASE and triplet loss, the model retains only one stream. The CNN takes the original \nnc{contour} sketch images as input and trains with a cross-entropy loss.
As shown in Table~\ref{table:ablation}, the performance degraded to 21.95\% at rank-1
accuracy in the cross-clothes matching, almost a 15\% lower matching rate than that of our full model.
This result indicates that performing deep convolution directly on \nnc{contour} sketch images is not effective for extracting reliable and discriminative features for cross-clothes person re-id.

\vspace{0.1cm}

\noindent \textbf{The performance of each stream.} In Table~\ref{table:stream},
we can see that the first stream outperformed the other streams because the range of \qqz{$\bm\theta$}
is the widest. However, \newc{when the number of sampled angles is fixed,} a wider range of \qqz{$\bm\theta$} means that the transformed image would
lose more details, \nncc{and that} is why the combination outperformed
the first stream. Moreover, because the images of stream 2 and stream 3
are transformed from different regions of the \nnc{contour} sketch image, their features are
complementary.

\subsection{\qz{Further Investigation of Our Model.}}\label{section:futher_evaluation}

\noindent \textbf{Analysis on the number of sampled angles and the learning strategy of the sampled angle.} 
\qzz{We varied the number of sampled angle, and the results are reported in Table~\ref{table:angle}. We can see that the performance of our model increases as the number of sampled angle increases; however, the computation would increase at the same time, \lastcc{and thus a trade-off is necessary}. The experimental results of different learning strategies of the sampled angle are reported in Table~\ref{table:angle}. \newc{This verified our analysis in Section~\ref{sec:lspt} that} if we update $\theta_i$ directly, 
\nnc{the range and the order of the sampled angle would be disrupted. For example, the pixels of $i$-th sampled angle of the transformed image are sampled from the region of foot while the pixels of the next sampled angle are sampled from the region of head. Thus, the semantic structure information of human is not preserved, which makes it difficult for CNN to learn discriminative feature. }
}

\begin{table}[t]
	\setlength{\tabcolsep}{12pt}
	\begin{center}
		\caption{\qzz{Performance (\%) on the number of sampled angles and the learning strategy of the sampled angle. For the first evaluaton, we vary the number of sampled angles. For the second evaluation, the ``Updating $\theta_i$ directly'' means we initialze $\theta_i$ from $\pi$ to $-\pi$ and update it by SGD directly instead of updating $\lambda_i$}.}
		\label{table:angle}
		\resizebox{\textwidth}{!}{%
			\begin{tabular}{|c||c|c|c|}

				\hline
				\multirow{2}{*}{\tabincell{c}{The number of\\ sampled angles}}
				&\multicolumn{3}{c|}{\tabincell{c}{Camera A and C (Cross-clothes)}} \\ \cline{2-4}

				                            & Rank 1         & Rank 10 & Rank 20 \\\hline
				112                    & 31.45 & 73.58   & 86.45   \\
				224                    & 34.38 & 77.30   & 88.05   \\
				336                    & 34.76          & 77.87   & 88.46   \\
				\hline
				\hline
				\multirow{2}{*}{\qzz{\tabincell{c}{The learning strategy\\ of the sampled angle}}}
				&\multicolumn{3}{c|}{\tabincell{c}{Camera A and C (Cross-clothes)}} \\ \cline{2-4}
				& Rank 1         & Rank 10 & Rank 20 \\\hline
				Updating $\theta_i$ directly   & 1.48 & 7.48   & 10.48   \\
				Updating $\lambda_k$ & 34.38 & 77.30   & 88.05   \\
				\hline
			\end{tabular}
		}
	\end{center}
\end{table}

\begin{figure}
	\centering
	\includegraphics[height=6cm,trim=0 0 0 0,clip]{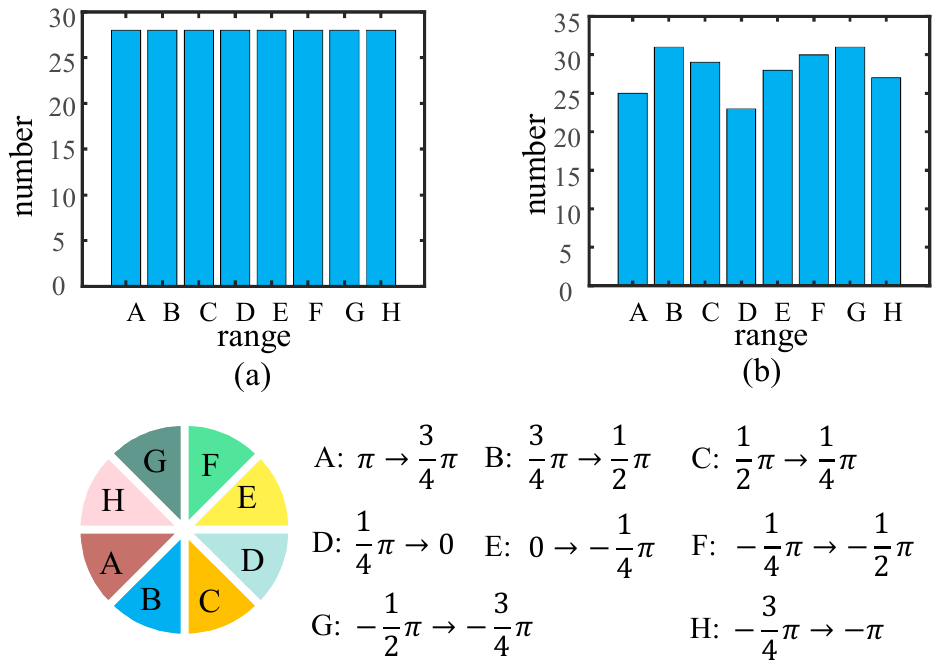}
	\caption{\qqz{The histograms of the $\bm{\theta}$. (a) The histogram of the $\bm{\theta}$ without training the SPT. (b) The histogram of the $\bm{\theta}$ with training the SPT. We can observe that the SPT samples more on the range of B, C, F and G from the original image.}}
	\label{fig:theta}
\end{figure}

\begin{figure*}[t]
	\centering
	\includegraphics[height=3cm,trim=0 400 0 50,clip]{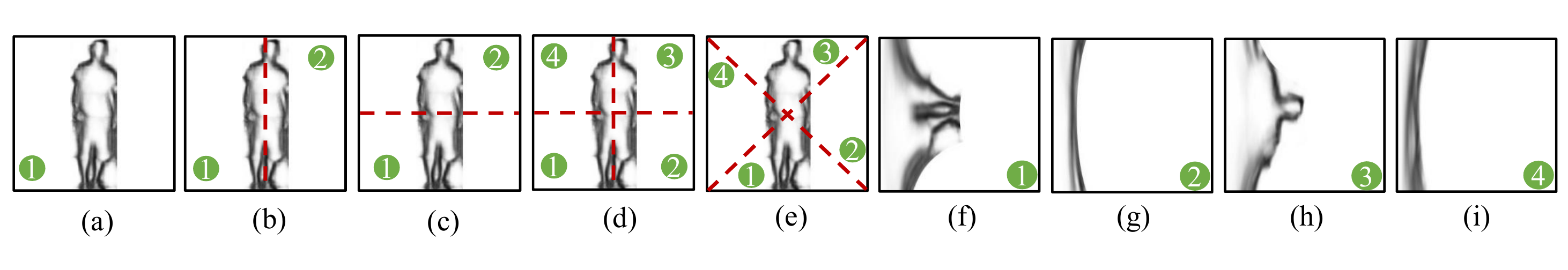}
	\caption{The different ranges of \qqz{$\bm\theta$}. (a), (b), (c), (d), (e)\nncc{:} Different partition
		strategies for an image; (f), (g), (h), (i)\nncc{:} the corresponding transformed
		images of (e) after different SPT layers. The numbers in green circles are the \ws{indeces}
		of the partitions, which are used in Table \ref{table:range}.}
	\label{fig:partition}
\end{figure*}

\begin{table*}[t]
	\setlength{\tabcolsep}{12pt}
	\begin{center}
		\caption{Analysis of different ranges of \qqz{$\bm\theta$} (the corresponding region is shown in
			Figure~\ref{table:range}), the performance (\%)
			is based on cosine similarity. In this experiment, a
			one-stream model is used for demonstration.}
		\label{table:range}
		\resizebox{\textwidth}{!}{%
			\begin{tabular}{l|c|c|c|c||c|c|c}
				\hline
				\multirow{2}{*}{\tabincell{c}{Range of $\bm\theta$ \\ \ \ (start $\rightarrow$ end)}}
				&\multirow{2}{3cm}{The corresponding \qzz{range} in Figure~\ref{fig:partition}}
				&\multicolumn{3}{l||}{Camera A and C (Cross-clothes)}
				&\multicolumn{3}{l}{Camera A and B (Same clothes)} \\ \cline{3-8}
				                                          &       & Rank 1 & Rank 10 & Rank 20 & Rank 1 & Rank 10 & Rank 20 \\\hline

				$\pi \qquad\, \rightarrow\  -\pi$         & (a).1 & 28.71  & 72.69   & 86.02   & 56.21  & 91.32   & 96.11   \\\hline

				$\pi \qquad\, \rightarrow\  0$            & (b).1 & 17.71  & 60.23   & 77.92   & 45.54  & 87.25   & 93.45   \\
				$0 \qquad\   \rightarrow\  -\pi$          & (b).2 & 19.11  & 54.71   & 72.21   & 33.21  & 78.23   & 90.12   \\\hline

				$-1/2\pi     \rightarrow\  -3/2\pi$       & (c).1 & 18.87  & 58.38   & 76.67   & 42.82  & 81.45   & 91.17   \\
				$1/2\pi \ \ \,   \rightarrow\  -1/2\pi$   & (c).2 & 20.64  & 62.78   & 77.45   & 45.78  & 83.58   & 92.84   \\\hline

				$\pi \qquad\   \rightarrow\  1/2\pi$      & (d).1 & 10.94  & 43.45   & 62.41   & 30.95  & 75.35   & 86.79   \\
				$1/2\pi \ \ \,     \rightarrow\  0$       & (d).2 & 13.35  & 49.45   & 65.94   & 25.99  & 67.21   & 83.42   \\
				$0 \qquad\;\,    \rightarrow\  -1/2\pi$   & (d).3 & 13.45  & 51.42   & 67.34   & 29.94  & 74.45   & 85.75   \\
				$-1/2\pi         \rightarrow\  -\pi$      & (d).4 & 10.89  & 43.78   & 61.51   & 28.15  & 71.86   & 84.46   \\\hline

				$3/4\pi \ \ \,      \rightarrow\  1/4\pi$ & (e).1 & 19.17  & 55.93   & 72.11   & 32.56  & 79.06   & 90.79   \\
				$1/4\pi \ \ \,  \rightarrow\  -1/4$       & (e).2 & 6.45   & 34.37   & 52.12   & 19.45  & 59.48   & 75.80   \\
				$-1/4\pi       \rightarrow\  -3/4\pi$     & (e).3 & 18.83  & 63.30   & 80.46   & 31.77  & 77.86   & 90.47   \\
				$-3/4\pi       \rightarrow\  -5/4\pi$     & (e).4 & 5.74   & 30.24   & 46.84   & 12.94  & 45.38   & 63.55   \\
				\hline

			\end{tabular}
		}
	\end{center}
\end{table*}

\vspace{0.1cm}

\noindent \textbf{Investigation of the range of \qqz{$\bm\theta$} (i.e., which part of the shape is important)}.
\qqz{We visualize the histogram of the $\bm\theta$ to demonstrate which part of the contour is more discriminative. As shown in Figure~\ref{fig:theta}, the SPT sampled more within the range of B, C, F and G from the original image and sampled less within the range of A and D. This indicates that the contours within range of B, C, F and G are more discriminative.}

We design different linear functions $f(z)$ to constrain \qqz{$\bm\theta$} learning within different ranges,
as shown in Table ~\ref{table:range}. The performance \ws{dropped} when \nncc{an image} was divided
into more partitions, not only for cross-clothes matching
but also for matching with no clothing change. Furthermore, different parts
performed differently. Take the partition strategy shown in
Figure~\ref{fig:partition} (e) as an example. The left and right parts
were less effective than the top and bottom parts because the transformed images of the left and the right parts are
straight lines with fewer details.
By contrast, the top and bottom parts
have more discriminating information, as we can see in Figure~\ref{fig:partition} (f)(h);
consequently, the performance is better.
Therefore, we can extract different \ws{local features by varying the constraint of \qqz{$\bm\theta$}.}


\begin{figure}
	\centering
	\includegraphics[height=5cm,trim=0 0 0 0,clip]{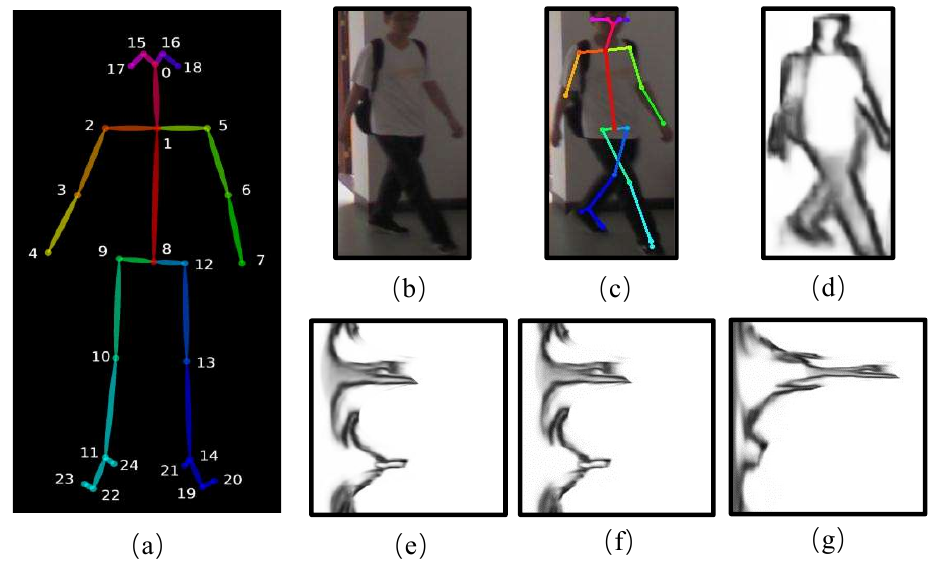}
	\caption{\revision{(a) The pose output format of BODY\_25 in ONENPOSE \cite{cao2018openpose,openpose_github}. (b) RGB image. (c) The estimated key-points. (d) Contour sketch image. (e) The transformed image by setting the center of the image as the transformation origin of SPT. (f) The transformed image by setting the middle of the hip \rchange{(i.e. the point of ``8'' in (a))} as the transformation origin of SPT. (g) The transformed image by setting the neck \rchange{(i.e. the point of ``1'' in (a))} as the transformation origin.}}
	\label{fig:skeleton}
\end{figure}

\begin{figure}
	\centering
	\includegraphics[height=8cm,trim=0 0 0 0,clip]{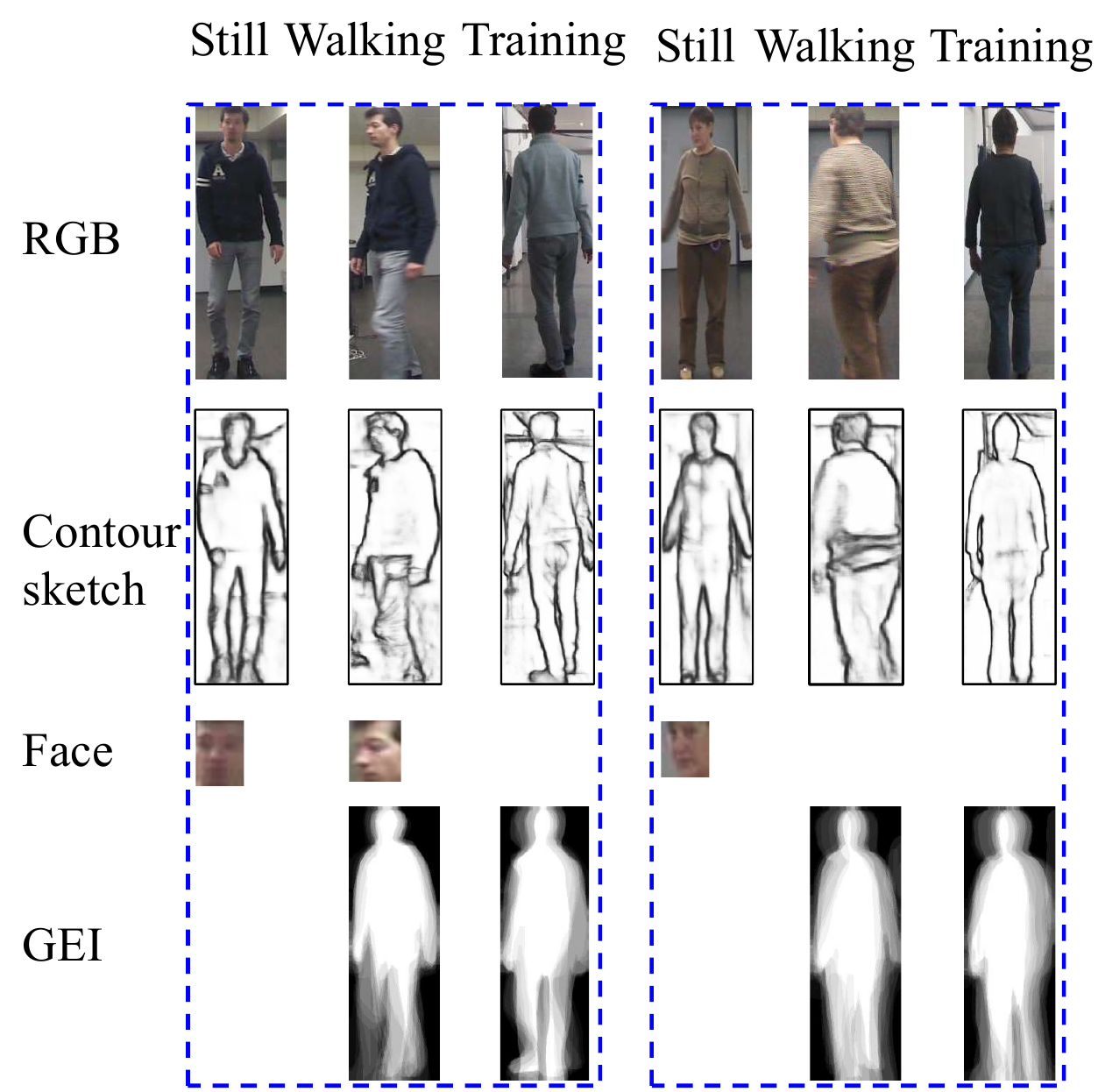}
	\caption{Examples of the BIWI dataset. The images in the first row are RGB images, and the following rows are \nnc{contour} sketch images, face images generated by face detector, and GEI images, \qzz{respectively. For each column, ``Still'', ``Walking'' and ``Training'' mean that the images are sample from the ``Still'', ``Walking'', ``Training'' subsets of BIWI, respectively.}
	\lastcc{Due to the viewpoint change, three examples have no observation of face; thus, the face detection method fails.}
	\nncc{Since the person is standing still in the ``Still'' subset, the GEI is not available.}
	\lastcc{The images in the same dash box are of the same identity.}}
	\label{fig:biwi}
\end{figure}

\begin{table}[t]
	
	\begin{center}
		\caption{\revision{The performances (\%) of different transformation origins used in our method on PRCC.}}
		\revision{
		\label{table:skeleton}
		\resizebox{\textwidth}{!}{%
			\begin{tabular}[t]{|c||l||c|c|c|}
				\hline
				Setting & The origin of SPT                                & Rank 1 & Rank 5 & Rank 10 \\ \hline
				\multirow{4}{*}{\tabincell{c}{Cross \\clothes}}
				        & Neck       & 32.93   &75.48    &  87.27   \\
				        & MiddleHip  & 32.35  &76.52	&88.43   \\
						&PTN\cite{esteves2018polar}	& 20.95	 & 65.85	& 81.39\\	
				        & The center of image (Ours)        & 34.38   &77.30	&88.05    \\
				\hline\hline
				\multirow{4}{*}{\tabincell{c}{Same \\clothes}}
				& Neck      &63.73	&94.20	&97.87   \\
				& MiddleHip  &67.03	&94.99	&98.14    \\
				&PTN\cite{esteves2018polar} & 36.21	& 81.86	& 91.87\\ 
				& The center of image (Ours)       &64.20	&92.62	&96.65    \\\hline
			\end{tabular}
		}}

	\end{center}

\end{table}

\begin{table}[t]
	\begin{center}
		\caption{The performance (\%) of our approach and the compared methods
			on the BIWI dataset. \qzz{The ``Still setting'' means the matching is between the subsets of ``Still'' and ``Training''; the ``Walking setting'' means the matching is between the subsets of ``Walking'' and ``Training''. Both settings are cross-clothes matching.}
		}
		\label{table:biwi}
		\resizebox{\textwidth}{!}{%
			\begin{tabular}[t]{|c||l||c|c|c|}
				\hline
				Setting & Methods                                & Rank 1 & Rank 5 & Rank 10 \\ \hline
				\multirow{7}{*}{\tabincell{c}{Still \\setting}}
				        & LOMO \cite{liao2015person}      & 6.68   & 43.97  & 89.31   \\
				        & LBP \cite{ojala1996comparative}  & 10.59  & 47.60  & 73.52   \\
				        & HOG \cite{dalal2005histograms}   & 9.73   & 35.91  & 76.59   \\\cline{2-5}
				        & PCB \cite{sun2017beyond} (RGB)         & 7.11   & 48.68  & 85.62   \\
						& Face \cite{wen2016discriminative}                & 15.66  & 42.31  & 74.36   \\
						
				        & Our model                              & \textbf{21.31}  & 66.10  & 90.02   \\
				\hline\hline
				\multirow{7}{*}{\tabincell{c}{Walking \\setting}}
				        & LOMO \cite{liao2015person}       & 3.14   & 33.24  & 83.87   \\
				        & LBP \cite{ojala1996comparative} & 9.28   & 45.43  & 76.95   \\
				        & HOG \cite{dalal2005histograms}   & 8.03   & 43.04  & 80.12   \\\cline{2-5}
				        & PCB \cite{sun2017beyond} (RGB)         & 6.49   & 47.05  & 89.60   \\
				        & Face \cite{wen2016discriminative}                & 12.80  & 40.56  & 73.50   \\
						& \revision{Gait \cite{han2006individual}}          & \revision{12.74}   & \revision{41.35}  & \revision{75.15}   \\
					
				        & Our model                              & \textbf{18.66}  & 63.88  & 91.47   \\\hline
			\end{tabular}
		}

	\end{center}
\end{table}

\vspace{0.1cm}

\noindent \revision{\textbf{Investigation of the transformation origin of SPT}.}
\revision{
	From another aspect, the SPT has the potential for combining with the key-point information of the body \cite{suh2018part, zhao2017spindle}. The SPT originally uses the center of contour sketch image as the transformation origin since we have a practical assumption that the center of the contour sketch image is close to the center of the human body for a detected person image. Alternatively, the SPT can also use pose key points as the transformation origin. Specifically, we use the OPENPOSE \cite{cao2018openpose} to estimate pose key points for each person image. We find that for most images, the estimations of the points of the neck and the middle of the hip are more precise as compared to other key points. Therefore, we use the points of the neck and the middle of the hip as the transformation origin of the SPT, respectively. If the pose estimation cannot detect these key points, we use the origin of the image as the transformation origin. By such an operation, the transformed images are shown in Figure~\ref{fig:skeleton}. 
}

\revision{
	We show the effect of the change on the operation of SPT in Table~\ref{table:skeleton}. It is found that using the center of the contour sketch image as the transformation origin of SPT is similar to the other operation\wss{s} but without more extra computation. It indicates that implementing our SPT using the center of sketch image as the origin is practical and acceptable.
	}
	
\begin{figure*}[t]
	\centering
	\includegraphics[height=4cm,trim=0 0 0 0,clip]{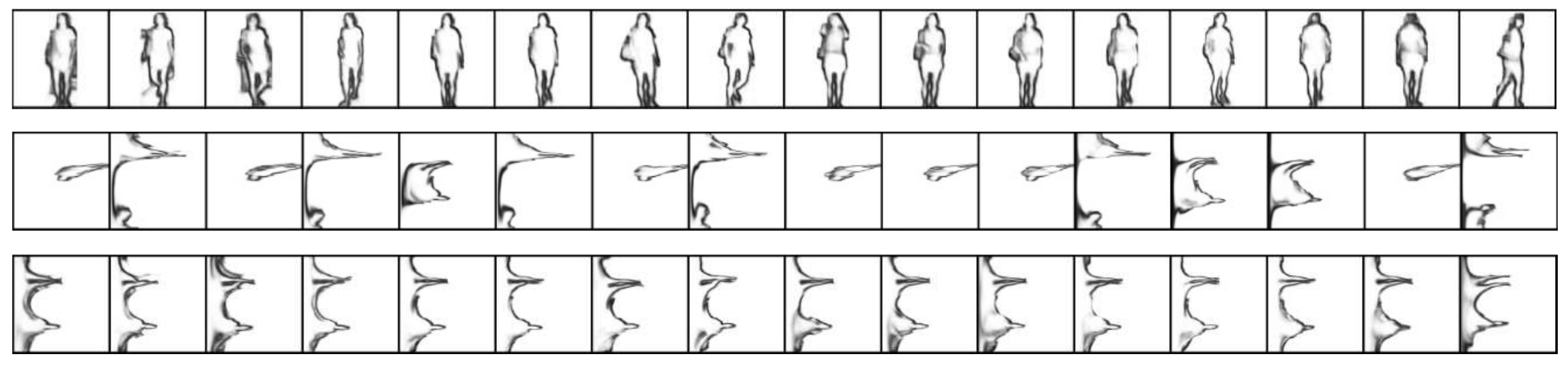}
	\caption{\revision{The images in the first row are contour sketch images of the same person in PRCC dataset after padding. The images in the second row are corresponding transformed images of the first row by using PTN in our method. The images in the third row are the corresponding transformed images of the first row by using the proposed SPT in our method.}}
	\label{fig:ptn}
\end{figure*}

\revision{
	In addition, we further conducted an experiment when replacing SPT with PTN in our method, and the results show that such a replacement degrades the performance \wss{as} shown in Table~\ref{table:skeleton}. And, a reason could be because PTN learns the transformation \wss{origin} for each image, \wss{and} it is shown in Figure \ref{fig:ptn} that different images of the same person have different transformation origins. \wss{T}herefore their transformed images are clearly different.
}

\subsection{Experiments on the BIWI Dataset}
The BIWI dataset \cite{munaro2014one} is a small publicly available dataset on clothing-change re-id, as shown in Figure~\ref{fig:biwi}. Due to the small size of the BIWI dataset, for comparison with the deep learning methods, we used our models that were pretrained on the PRCC dataset and fine-tuned them on the training set. The BIWI dataset was randomly split in half for training and testing without overlapping identities. 

The experimental results on the BIWI dataset are reported in Table~\ref{table:biwi}.
\qqz{Our method clearly outperformed the hand-crafted features and PCB under the still setting as well as under the walking setting.} 
\revision{Since video segments are available for BIWI, similar to the experiments in \cite{wang2016person} that compares re-id methods with gait recognition methods, we used DeepLabV3 \cite{chen2017rethinking} to generate gait silhouettes for each sequence. We computed the gait energy image (GEI) for each image sequence and used the DGEI method \cite{han2006individual} for matching. In the ``Still'' subset, the person is standing still, so we do not compare gait recognition in this scenario. However, one challenge in applying gait recognition to unregulated image sequences in re-id is that the sequence must include a complete normal gait cycle.}

In addition, we wish to see how face recognition performs at a distance in a clothing-change scenario. We compared our method with \nncc{the} face recognition method \lastcc{\cite{wen2016discriminative}} \nncc{again} and found that the face recognition method outperformed gait recognition and RGB-based person re-id but not our proposed method. The \nncc{inferior} performance occured because of the occlusion, viewpoint changes and low resolution \newc{(the average resolution of the detected faces is about \lastcc{34/27} pixels in height/width)} of person images \ws{as well as false face detection}. Furthermore, in the BIWI dataset, the face images in the ``Still'' subset are better observed than those in the ``Training'' and ``Walking'' subsets, so the face recognition method performs better under the ``Still'' setting. Hence, whether a face is better observed limits the performance of face recognition in surveillance; \lastcc{for instance, when there is no observation of face on the back of a person, the matching associated to this person image using face recognition will be failed.}

\begin{table}[t]
	\begin{center}
		\caption{\revision{The performances (\%) of our proposed method and compared methods under cross-dataset evaluation setting. ``RGB'' means the inputs are RGB images.
		}}
		\label{table:cross}
       
		\resizebox{\textwidth}{!}{%
			\begin{tabular}[t]{|c||l||c|c|c|}
				\hline
				Setting & Methods                                & Rank 1 & Rank 5 & mAP \\ \hline	
				\multirow{3}{*}{\tabincell{c}{\roundtwo{Duke \cite{ristani2016performance,zheng2017unlabeled}} to \\ Market \cite{zheng2015scalable}}}
						& Res-Net50 \cite{he2016deep} (RGB)   & 56.26  & 71.35  & 24.52   \\	
						& Ours             & 26.19  & 44.45  & 21.70   \\
            & Ours + Res-Net50 \cite{he2016deep} (RGB)             & 59.23  & 73.52  & 26.27   \\ \hline\hline
        \multirow{3}{*}{\tabincell{c}{Market \cite{zheng2015scalable} \\ to \roundtwo{Duke \cite{ristani2016performance,zheng2017unlabeled}}}}
            & Res-Net50 \cite{he2016deep} (RGB)  & 27.11  & 40.62  & 14.00   \\	
						& Ours         & 23.52  & 35.46  & 7.74   \\
						& Ours + Res-Net50 \cite{he2016deep} (RGB)            & 35.10  & 48.07  & 17.40   \\\hline
           
			\end{tabular}
		} 
	\end{center}
\end{table}

\subsection{\revision{Experiments on Cross-Dataset Evaluation}}
\revision{
To show the generalizability of our model for cross-dataset person re-identification, another challenge on conventional against intra-dataset person re-identification, we trained our model on the training set of Market-1501 or DukeMTMC-reID dataset and then tested it on the testing set of another dataset (See supplementary for the use of our contour sketch feature on conventional person re-identification). The results are reported in Table~\ref{table:cross}. As such, since the color appearance is important for conventional person re-identification with no clothing change problem, the combination of our method and Res-Net50 (RGB) achieves the best performance on both datasets, where we weight the distance of our method by 0.7 and the one of Res-Net50 (RGB) by 1. This experiment further shows the potential use of our proposed contour-sketch-based method.}


\section{Conclusion}

We have attempted to address the challenging problem of clothing changes in person re-id using visible light images.
In this work, we have proposed the extraction of discriminative features from contour sketch images to address moderate clothing changes. \ws{We assume that a person wears clothes of similar thickness and}, thus the shape of a person would not change significantly when the weather does not change substantially within a short period of time.


To extract reliable and discriminative curve patterns from the contour sketch, we introduce a \newc{learnable} spatial polar transformation (SPT) \newc{into} the deep neural network for selecting discriminant curve patterns and an angle-specific specific (ASE) for mining fine-grained \qz{angle}-specific features.
We then form a multistream deep learning framework by varying the sampling range of the SPT to aggregate \newc{multi-granularity (i.e. coarse-grained and fine-grained)} features.
Extensive experiments conducted on our \newc{developed} re-id dataset and the existing BIWI dataset validate the effectiveness and stability of our contour-sketch-based method on cross-clothes matching compared with RGB-based methods.
Our experiments also show the challenge of cross-clothes person re-id, which became intractable when a clothing change is combined with other large variations.

While our attempt has shown that contour sketches can be used to solve the cross-clothes person re-id problem, we believe that multimodal learning incorporating nonvisual cues is another potential approach to solve this problem.

\ifCLASSOPTIONcompsoc
	\section*{Acknowledgments}
\else
	\section*{Acknowledgment}
\fi

This work was supported partially by the National Key Research and Development Program of China  (2016YFB1001002), NSFC (U1811461), Guangdong Province Science and Technology Innovation Leading Talents (2016TX03X157), and Guangzhou Research Project (201902010037). The corresponding author for this paper and the principal investigator for this project is Wei-Shi Zheng.

\ifCLASSOPTIONcaptionsoff
	\newpage
\fi



\bibliographystyle{IEEEtran}
\bibliography{IEEEabrv,./paper}
%



%
\begin{IEEEbiography}[{\includegraphics[width=1in,height=1.25in,clip,keepaspectratio]{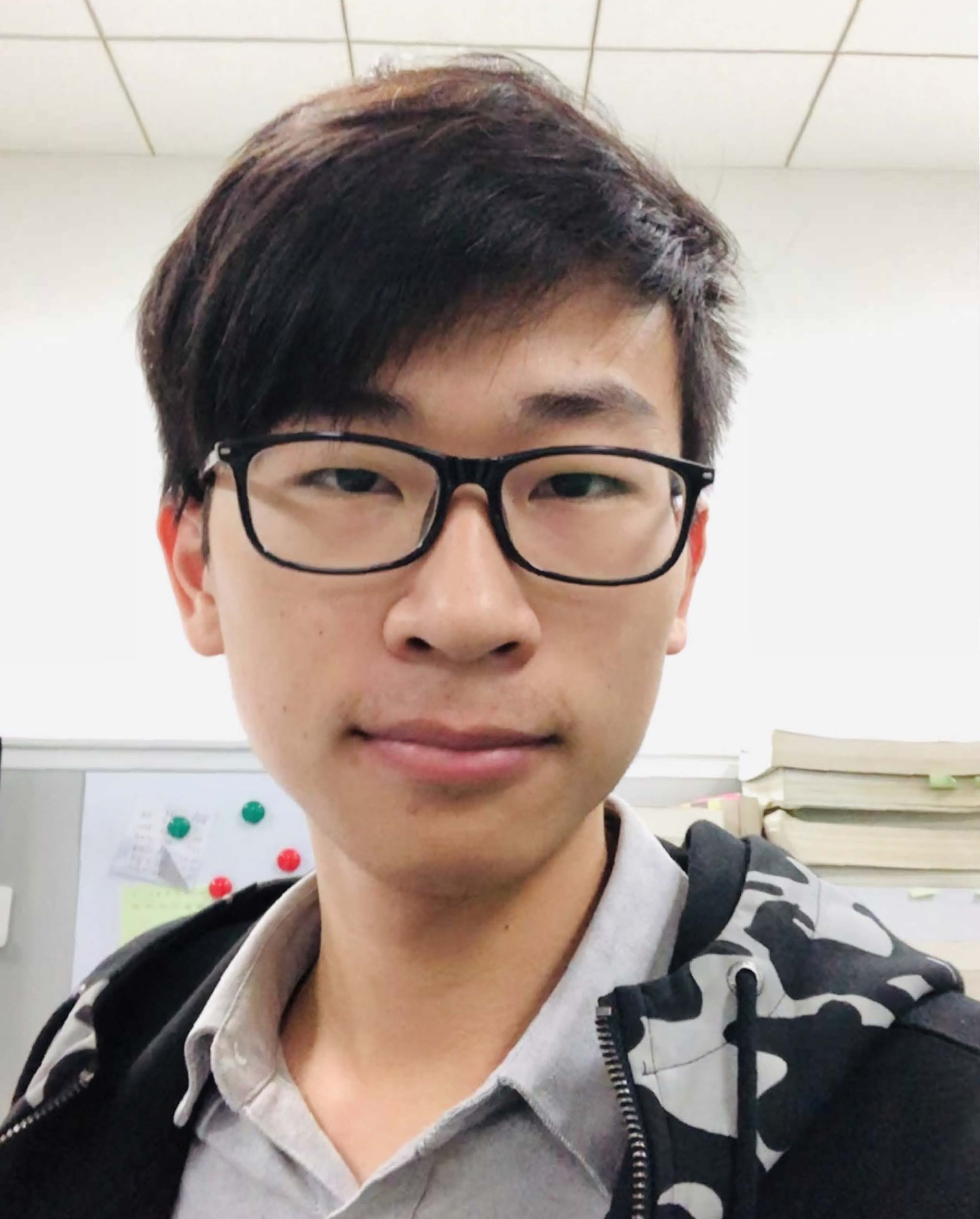}}]{Qize Yang} received Bachelor's degree in information engineering from South China University of Technology in 2017. He is now an M.S. student in the School of Data and Computer Science at Sun Yat-sen University. His research interest lies in computer vision and machine learning.
\end{IEEEbiography}

\begin{IEEEbiography}[{\includegraphics[width=1in,height=1.25in,clip,keepaspectratio]{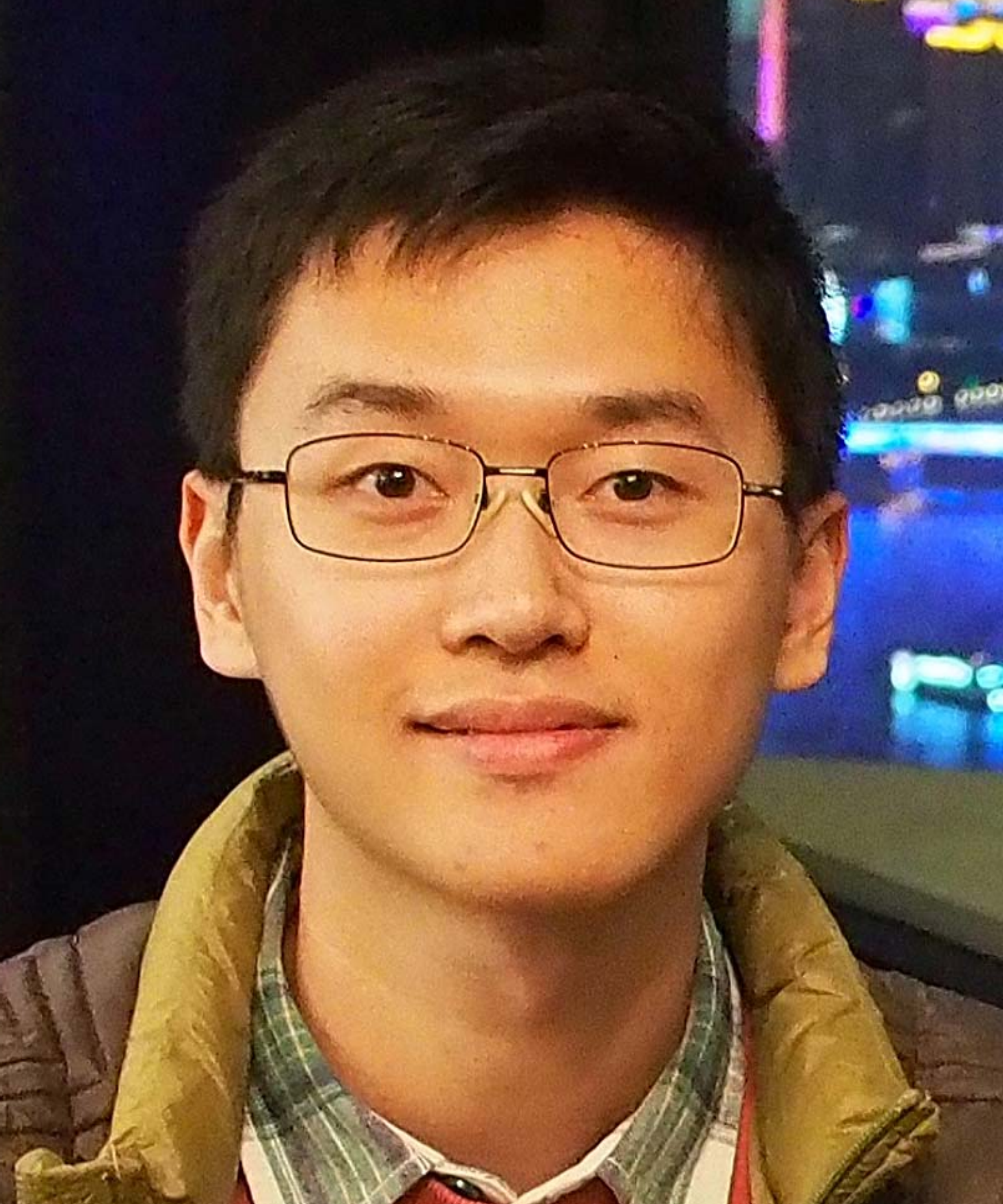}}]{Ancong Wu} received a Bachelor's degree in intelligence science and technology from Sun Yat-sen University in 2015. He is pursuing a PhD degree with the School of Electronics and Information Technology at Sun Yat-sen University. His research interests are computer vision and machine learning. He is currently focusing on the topic of person re-identification.
\end{IEEEbiography}

\begin{IEEEbiography}[{\includegraphics[width=1in,height=1.25in,clip,keepaspectratio]{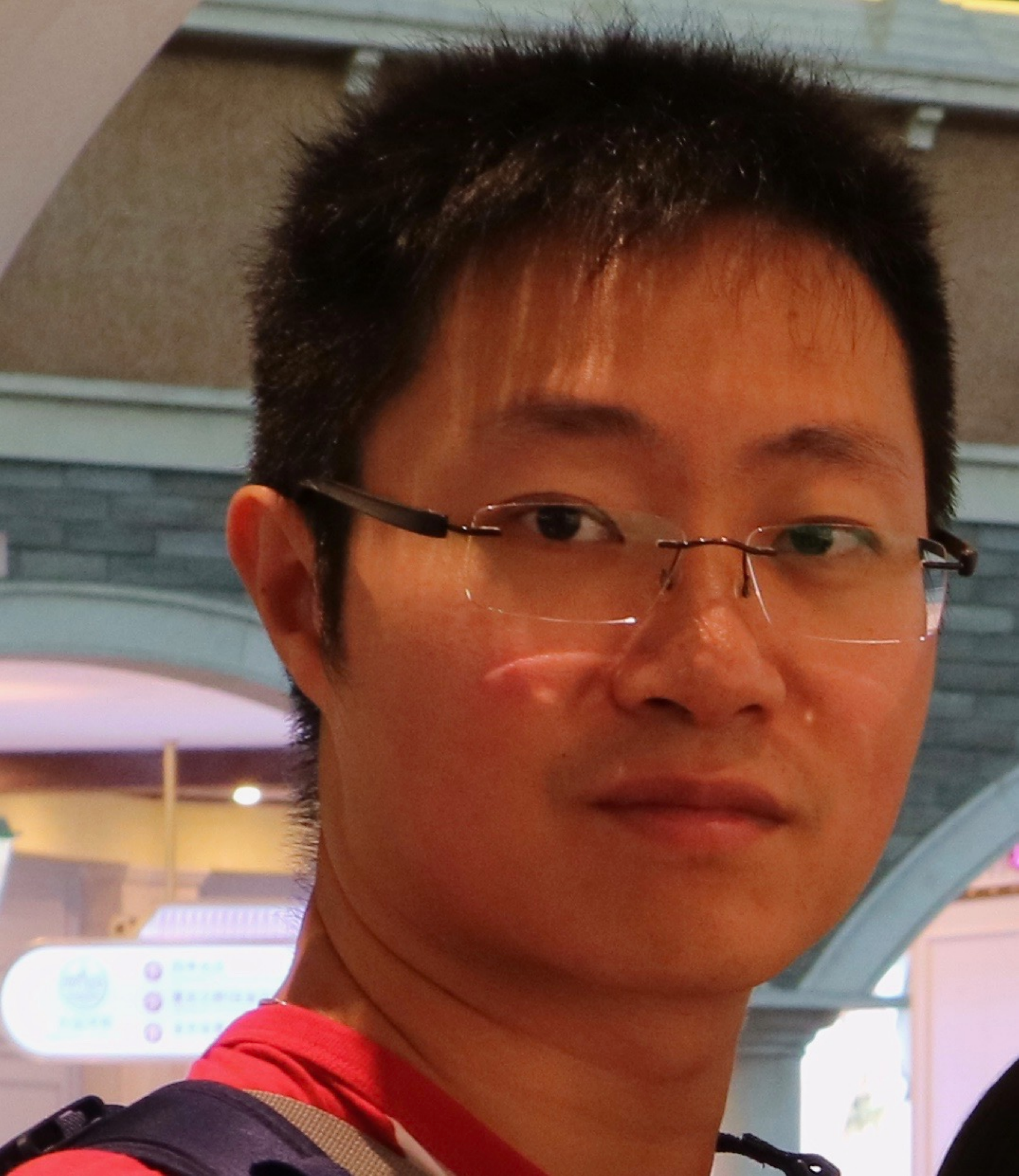}}]{Wei-Shi Zheng} is a professor at Sun Yat-sen University. His research interests include person association and activity understanding in visual surveillance. He has published more than 120 papers, including more than 90 publications in major journals (IEEE Transactions on Pattern Analysis and Machine Intelligence, IEEE Transactions on Neural Networks and Learning Systems, IEEE Transactions on Image Processing, Pattern Recognition) and top conferences (ICCV, CVPR,IJCAI,AAAI). He served as an area chair/SPC for AVSS 2012, ICPR 2018, BMVC 2018/2019, IJCAI 2019/2020 and AAAI 2020. He has joined Microsoft Research Asia Young Faculty Visiting Programme. He is a recipient of Excellent Young Scientists Fund of the National Natural Science Foundation of China, and is a recipient of the Royal Society-Newton Advanced Fellowship, United Kingdom. He is an associate editor of the Pattern Recognition journal.
\end{IEEEbiography}
\end{document}